\documentclass[conference]{IEEEtran}
\IEEEoverridecommandlockouts
\usepackage{threeparttable}
\usepackage{times}
\usepackage{soul}
\usepackage{url}
\usepackage[hidelinks]{hyperref}
\usepackage[utf8]{inputenc}
\usepackage[small]{caption}
\usepackage{graphicx}
\usepackage{amsthm}
\usepackage{booktabs}
\usepackage{algorithm}
\usepackage{algorithmic}
\usepackage{subfigure}
\usepackage{amssymb}
\usepackage{xcolor}
\def\model{GAINS}

\usepackage{cite}
\usepackage{amsmath,amssymb,amsfonts}
\usepackage{algorithmic}
\usepackage{graphicx}
\usepackage{textcomp}
\usepackage{xcolor}
\def\BibTeX{{\rm B\kern-.05em{\sc i\kern-.025em b}\kern-.08em
    T\kern-.1667em\lower.7ex\hbox{E}\kern-.125emX}}
\begin{document}

\title{Beyond Discrete Selection: Continuous Embedding Space Optimization for Generative Feature Selection
}

\author{\IEEEauthorblockN{Meng Xiao$^{1,2,\S}$, Dongjie Wang$^{3,\S}$\thanks{$\S$ These authors have contributed equally to this work.}, Min Wu$^4$, Pengfei Wang$^{1,2}$, Yuanchun Zhou$^{1,2}$, Yanjie Fu$^{5,*}$\thanks{$^*$ Corresponding author.}}
\IEEEauthorblockA{\textit{$^1$Computer Network Information Center, Chinese Academy of Sciences, China} \\
\textit{$^2$University of Chinese Academy of Sciences, China} \\
\textit{$^3$Department of Computer Science, University of Central Florida, United States}\\
\textit{$^4$Institute for Infocomm Research (I$^2$R), Agency for Science, Technology and Research (A*STAR), Singapore}\\
\textit{$^5$Arizona State University, School of Computing and AI, United States}\\
\thanks{The first author completed this work while serving as a visiting student at Institute for Infocomm Research (I$^2$R), Agency for Science, Technology and Research (A*STAR), Singapore.}
shaow@cnic.cn, wangdongjie@knights.ucf.edu, wumin@i2r.a-star.edu.sg, \{wpf, zyc\}@cnic.cn, yanjiefu@asu.edu}}


\maketitle

\begin{abstract}
The goal of Feature Selection - comprising filter, wrapper, and embedded approaches - is to find the optimal feature subset for designated downstream tasks. Nevertheless, current feature selection methods are limited by: 1) the selection criteria of these methods are varied for different domains, leading them hard to be generalized; 2) the selection performance of these approaches drops significantly when processing high-dimensional feature space coupled with small sample size. In light of these challenges, we pose the question: can selected feature subsets be more robust, accurate, and input dimensionality agnostic?
In this paper, we reformulate the feature selection problem as a deep differentiable optimization task and propose a new research perspective: conceptualizing discrete feature subsetting as continuous embedding space optimization. 
We introduce a novel and principled framework that encompasses a sequential encoder, an accuracy evaluator, a sequential decoder, and a gradient ascent optimizer. This comprehensive framework includes four important steps: preparation of features-accuracy training data, deep feature subset embedding, gradient-optimized search, and feature subset reconstruction.
Specifically, we utilize reinforcement feature selection learning to generate diverse and high-quality training data and enhance generalization. By optimizing reconstruction and accuracy losses, we embed feature selection knowledge into a continuous space using an encoder-evaluator-decoder model structure. We employ a gradient ascent search algorithm to find better embeddings in the learned embedding space. Furthermore, we reconstruct feature selection solutions using these embeddings and select the feature subset with the highest performance for downstream tasks as the optimal subset. 
Finally, extensive experimental results demonstrate the effectiveness of our proposed method, showcasing significant enhancements in feature selection robustness and accuracy. To improve the reproducibility of our research, we have released accompanying code and datasets by Dropbox.
\footnote{\tiny \url{https://www.dropbox.com/sh/vlwz16cquv5ct9d/AACvhlDRwBe3f4nWwGEE4zeCa?dl=0}}.

\end{abstract}


\section{Introduction}
Feature selection aims to identify the most appropriate subset of features, which can be employed to optimize downstream predictive tasks.
A proficiently executed feature selection can  reduce dimensionality, shorten training time, bolster the generalization capability, mitigate the risk of overfitting,  enhance predictive accuracy, and improve interpretation and explanation. 
Thus, feature selection can enrich our understanding of the underlying data patterns for conducting more comprehensive analyses.

Within the scope of practical application, feature selection encounters two prominent challenges: 1) The aspect of generalization, and 2) The issue of robustness.
Firstly, different selection algorithms use different criteria to select representative features, making it challenging to find the best algorithm for cross-domain datasets. The notion of generalization within feature selection aims to address the following question: how can we facilitate consistently high accuracy across multiple domains?
Secondly, certain domains (e.g., biomedical)  involve a huge number of features, but sample sizes are limited due to costs, privacy, and ethnicity. 
When data are high-dimensional, point-point distances tend to be the same, and data patterns are non-discriminative. Concurrently, high dimensionality can amplify  feature selection complexity and time costs in a discrete space. On the other hand, when data are low sample-sized,  distributions are sparse, and data patterns are unclear.
Addressing robustness within feature selection intends to answer the following question: how can we automatically identify an effective yet small-sized feature subset with input dimensionality-agnostic time costs?

Relevant studies can only partially solve the two challenges. 
Classic feature selection algorithms can be grouped into three categories: 
(i) filter methods (e.g., univariate feature selection \cite{kbest,forman2003extensive}, correlation-based feature selection \cite{hall1999feature,yu2003feature}), in which features are ranked by a specific score. 
However, feature relevance or redundancy scores are usually domain-specific, non-learnable, and cannot generalize well to all applications. 
(ii) wrapper methods (e.g., evolutionary algorithms \cite{yang1998feature,kim2000feature}, branch and bound algorithms \cite{narendra1977branch,kohavi1997wrappers}), in which optimal feature subset is identified by a search strategy that collaborates with predictive tasks. 
However, such methods have to search a large feature space of  $2^N$ feature subspace candidates, where $N$ is the number of input features. 
(iii) embedded methods (e.g., LASSO \cite{lasso}, decision tree \cite{sugumaran2007feature}), in which feature selection is part of the optimization objective of predictive tasks.
However, such methods are subject to the strong structured assumptions of predictive models (e.g., the L1 norm assumption of LASSO). That is, feature selection and predictive models are coupled together, lack flexibility, and are hard to generalize to other predictors. 
Therefore, we need a novel perspective to derive the novel formulation and solver of generalized and disruption-robust feature selection.


\textbf{Our contributions: a discrete subsetting as continuous optimization perspective.}
We formulate the problem of discrete feature selection as a gradient-based continuous optimization task. 
We propose a new perspective: the discrete decisions (e.g., select or deselect) of feature selection can be embedded into a continuous embedding space, thereafter, solved by a more effective gradient-based solution. 
We show that this perspective can be implemented by feature subset encoding, gradient-optimized search, and reconstruction. 
Training deep models requires big training data, and we find that reinforcement feature selection can be used as a tool to automatically generate features-accuracy pairs in order to increase the automation, diversity, and scale of training data preparation. 
We demonstrate that integrating both reinforcement-generated and classic selection algorithms-generated experiences learn a better feature subset embedding space because: 1) using reinforcement is to crowd-source unknown exploratory knowledge and 2) using classic selection algorithms is to exploit existing peer knowledge.
We highlight that a globally-described discriminative embedding space, along with the joint objective of minimizing feature subset reconstruction loss and accuracy estimation loss,  can strengthen the denoising ability of gradient search, eliminate noisy and redundant features, and yield an effective feature subset. 
We observe that, by representing feature subsets into fixed-sized embedding vectors, the time costs of gradient-based optimization are input dimensionality-agnostic (only relates to embedding dimensionality). 
The feature subset decoder can automatically reconstruct optimal selected features and doesn't need to manually identify the number of best features.

\textbf{Summary of Proposed Approach.} Inspired by these findings, this paper presents a generic and principled framework for deep differentiable feature selection. 
It has two goals: 1) generalized across domains; 2) disruption-robust: overcome training data bottlenecks, reduce feature subset size while maintaining accuracy, and control time costs against input dimensionality. 
To achieve Goal 1,  we achieve feature selection via a pipeline of representation-search-reconstruction. 
Specifically, given a downstream predictor (e.g., decision tree), we collect the features-accuracy pairs from diverse feature selection algorithms so that training data represent the diversity of samples.
An embedding space is learned to map feature subsets into vectors. 
We advance the representability and generalization of the embedding space by exploiting the diverse and representative train samples to optimize the joint loss of feature subset reconstruction and accuracy evaluation.
We leverage the accuracy evaluator as feedback to infer gradient direction and degree to improve the search for optimal feature subset embeddings. 
To achieve Goal 2, we devise different computing strategies: 1) We develop a multi-agent reinforcement feature selection system that can automatically generate large-scale high-quality features-accuracy pairs in a self-optimizing fashion and overcome training data bottlenecks. 
2) We reformulate a feature subset as an alphabetical non-ordinal sequence. We devise the strategy of jointly minimizing not just accuracy evaluation loss but also sequence reconstruction loss in order to position the gradient toward a more denoising direction over the continuous embedding space to reconstruct a shorter feature subset. 
We find that the reduction of feature space size improves generalization.
3) Since we embed feature subsets into a fixed-size embedding space, gradient search in a fixed-size space is input dimensionality agnostic. 
Finally, we present extensive experimental results to show the small-sized, accurate,  input-dimensionality agnostic, and generalized properties of our selected features.

\section{Problem Statement}

Our goal is to develop a generalized and robust deep differentiable FS framework.
Formally, given a dataset $D=\{X, y\}$, where $X$ is a feature set, and $y$ is predictive target labels. 
We utilize classic FS algorithms to $D$ to collect $n$ feature subset-accuracy pairs as training data, denoted by $R=\{(\mathbf{f}_i, v_i)\}_{i=1}^n$, where $\mathbf{f}_i=[f_1,f_2,\cdots,f_T]$ is the feature ID token set of the $i$-th feature subset, and $v_i$ is corresponding downstream predictive accuracy. 
Thereafter, we pursue two aims: 
1) constructing an optimal continuous space of feature subsets. 
We learn a mapping function $\phi$, a reconstructing function $\psi$, and an evaluation function $\omega$ via joint optimization to convert $R$ into a continuous embedding space $\mathcal{E}$, in which each embedding vector represents the feature ID token set of a feature subset and corresponding model performance.
2) searching the optimal feature subset.
We adopt gradient-based search to find the optimal feature token set $\mathbf{f}^{*}$, given by: 
\begin{equation}
    \mathbf{f}^{*} = \psi(\mathbf{E}^{*}) = \text{argmax}_{\mathbf{E}\in\mathcal{E}} \mathcal{A}(\mathcal{M}(X[\psi(\mathbf{E})]),y), 
\end{equation}
where $\psi$ is a reconstruction function to reconstruct a feature ID token set from any embedding of $\mathcal{E}$;
$\mathbf{E}$ is an embedding vector in $\mathcal{E}$ and $\mathbf{E}^{*}$ is the optimal one;
$\mathcal{M}$ is the downstream ML model and $\mathcal{A}$ is the performance indicator.
We apply $\mathbf{f}^{*}$ to $X$ to select the optimal feature subset $X[\mathbf{f}^*]$ to maximize  downstream ML model performances.

\section{Deep Differentiable Feature Selection}
In this section, we present an overview of our method, then introduce the technical details of each component. 
 
\begin{figure*}[!h]
\centering
\includegraphics[width=\linewidth]{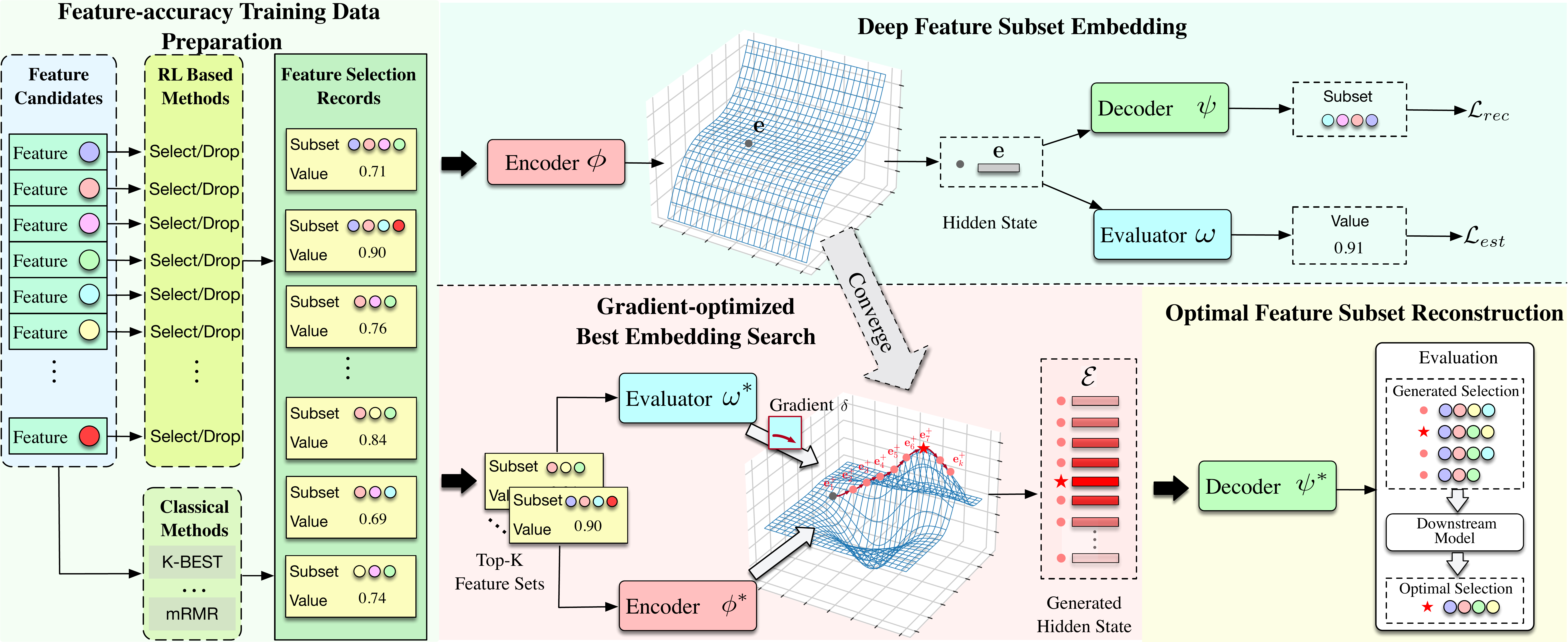}
\caption{An overview of our framework. \model\ is made up of four main components: 1) feature-accuracy training data preparation, designed to quickly collect large quantities of qualified and valid training data; 2) deep feature subset embedding, purposed to preserve the knowledge of feature selection into a global continuous embedding space; 3) gradient-optimized best embedding search, intended to search for better embeddings in the learned space; 4) optimal feature subset reconstruction, devised to reconstruct and output the optimal feature subset.}
\label{model_overview}
\end{figure*}

\subsection{Framework Overview}
Figure~\ref{model_overview} shows the overview of our framework, including four steps: 1) feature-accuracy training data preparation; 2) deep feature subset embedding; 3) gradient-optimized best embedding search; 4) optimal feature subset reconstruction. 
Step 1 is to collect selected feature subsets and corresponding predictive performances as training data for deep feature subset embedding.
In particular, to exploit existing peer knowledge, we utilize classical feature selection methods (e.g., K-Best, mRMR) to collect feature subset-accuracy records;
to explore crowdsource unknown knowledge, we utilize reinforcement learning (RL) based feature selector to collect diverse feature subset-accuracy records.
Step 2 is to develop a deep encoder-evaluator-decoder model to learn the optimal embedding space of feature subsets.
In particular, given a feature subset, the encoder maps the feature subset ID tokens into a continuous embedding vector; 
the evaluator optimizes the embedding vector along the gradient direction by predicting the corresponding model performance; 
the decoder reconstructs the feature ID tokens using feature subset embedding vectors. 
We learn the optimal embedding space by jointly optimizing the reconstruction and evaluation losses of the encoder, evaluator, and decoder. 
Step 3 aims to expedite the gradient-optimized search of optimal feature subset embedding vector in the embedding space. 
In particular, we first select top-K historical feature subset-accuracy pairs as search seeds (starting points) and obtain corresponding embeddings using the well-trained encoder.
We then search by starting from these embeddings at a minute rate in the gradient direction of performance improvement to identify optimal embeddings.
Step 4 is to exploit the well-trained decoder to reconstruct the optimal feature ID tokens as candidate feature subsets from these identified embeddings.
We evaluate these candidate feature subsets with a downstream ML model to present the best feature subset with the highest performance.

\subsection{Feature-Accuracy Training Data Preparation}
We collect a set of feature-accuracy pairs as training data to construct an effective continuous space that depicts the properties of the original feature set.

\smallskip
\noindent\textbf{Leveraging exploitation and exploration for automated training data preparation.}
The diversity, scale, and comprehensiveness of the training data population are essential for learning an effective embedding space. 
The training data is collected via two strategies. 1) \textit{exploitation of existing feature selection algorithms}: we apply classical feature selectors (e.g., KBest, Lasso.) to the given feature set. Each algorithm represents one algorithmic perspective and produces a small number of feature subset-accuracy records. It is challenging to collect large-scale, diverse training data. 
2) \textit{exploration of other unknown candidate subsets}: We find that reinforcement learning can be used to automatically explore and select various feature subsets to evaluate corresponding feature subset performance~\cite{marlfs}. In other words, reinforcement feature selection can be viewed as a tool for automated training data preparation.  
In particular, we develop a multi-agent reinforcement feature selection system, where each agent is to select or deselect a feature in order to progressively explore feature subsets and find high-accuracy low-redundancy feature subsets. The reinforcement feature selection system leverage randomness and self-optimization to automatically generate high-quality, diverse, comprehensive feature subset-accuracy records to overcome data sparsity.

\smallskip
\noindent\textbf{Leveraging order agnostic property to augment training data.}
To learn a feature subset embedding space, we view a feature subset as a token sequence and exploit seq2vec models to embed a feature subset as a vector. 
Since the feature subset is order-agnostic, the token sequence is order-agnostic. 
We leverage the order-agnostic property to augment the feature subset-accuracy training data by shuffling and rearranging the orders of selected features in a feature token sequence.
Formally, given a selected feature subset and corresponding predictive accuracy, denoted by $\{\mathbf{f},v\}$, where $\mathbf{f} = [f_1,f_2,\cdots,f_T]$ is the ID tokens of selected features and $v_i$ is corresponding performance.
We shuffle the old token order and obtain a new order of the $T$-length feature subset $\mathbf{\tilde{f}}=[f_3,f_T,\cdots,f_1]$.
We add the new shuffled token sequence and corresponding accuracy into the training data to enhance data diversity and comprehensiveness.

\subsection{Deep Feature Subset Embedding}
\noindent\textbf{Why effective embedding space construction matters.} 
Most conventional feature selection algorithms are of a discrete choice formulation.
Such formulation results in suboptimal performance since it is difficult to enumerate all possible feature combinations in a high-dimensional dataset.
To efficiently search the optimal feature subset, it is crucial to change feature selection from searching in discrete space to searching in continuous space, where gradient-based optimization can be used for faster and more effective selection.

But, How can the knowledge derived from feature selection be incorporated into an embedding space? There are two potential methods: 1) Sequential modeling, which learns distinct embeddings for the same feature subset irrespective of order; 2) Set modeling, which generates a consistent embedding for the same feature subset, even with altered order. 
In sequential modeling, the identical subset with varying orders yields different embeddings. These form a global continuous embedding space, where lighter regions signify poor model performance, while darker regions correspond to superior performance. It is easy to search for the optimal solution by shifting candidate embeddings to the optimal regions via gradients.
Conversely, set modeling, which learns a consistent embedding for variations of the same feature subset, results in a locally optimal embedding space. This model cannot perceive the order of feature subsets that do not influence performance. Consequently, even after embedding updates, locating the optimal point remains challenging due to the lack of clear direction.
Based on these observations, we opt for sequential modeling (i.e., LSTM) to integrate feature selection knowledge into a continuous embedding space.

\smallskip
\noindent\textbf{Leveraging the encoder-evaluator-decoder framework to embed feature subset tokens.}
We aim to construct an effective continuous embedding space in order to search for better feature subsets using gradient-based search in the gradient direction of higher performance.
We develop a novel learning framework that includes the encoder, evaluator, and decoder.
Next, we use $\mathbf{f}$ to denote the feature ID tokens of selected features and $v$ to denote the corresponding model performance.

\smallskip
\noindent\textbf{The feature subset encoder $\phi$:}
The encoder is to learn a mapping function $\phi$ that takes  $\mathbf{f}$ as input, and outputs its continuous embedding $\mathbf{E}$, denoted by $\phi(\mathbf{f}) = \mathbf{E}$.
The encoder is designed based on a single layer of Long Sort-Term Memory~\cite{lstm} (LSTM), where $\mathbf{f} \in\mathbb{R}^{1\times T}$ and  $T$ is the length of the feature subset. 
After inputting $\mathbf{f}$ into $\phi$, we collect each feature ID token embedding to form $\mathbf{E}$. 
Here, $\mathbf{E} = [\mathbf{h}_1, \mathbf{h}_2,\cdots, \mathbf{h}_T] \in \mathbb{R}^{T\times d}$, where $\mathbf{h}_t \in \mathbb{R}^{1\times d}$ is  the $t$-th token embedding with dimension  $d$.


\smallskip
\noindent\textbf{The feature subset decoder $\psi$:}
The decoder $\psi$ aims to reconstruct the feature ID tokens  $\mathbf{f}$ of of an embedding vector $\mathbf{E}$, denoted by $\psi(\mathbf{E})=\mathbf{f}$. 
Similar to the encoder, we employ a single-layer LSTM to implement the decoder.
We train the decoder in an autoregressive manner.
The initial input of the decoder is the last embedding of $\mathbf{E}$, denoted by $\mathbf{\hat{h}}_0=\mathbf{h}_T$.
We take the $i$-th step in LSTM as an example to demonstrate the calculation process.
Specifically, we input $\mathbf{\hat{h}}_{i-1}$ and the $i$-th feature ID token in $\mathbf{s}$ into the LSTM layer to get the current embedding $\mathbf{\hat{h}}_i$.
The output of the decoder and the input of the encoder share the same feature ID token dictionary.
Thus, to forecast the next most possible token easily, we utilize the attention mechanism~\cite{attention} to learn the attention weights between  $\mathbf{\hat{h}}_i$ and each token embedding in $\mathbf{E}$.
Then, we generate the enhanced embedding $\mathbf{\tilde{h}}_i$ by aggregating the knowledge of $\mathbf{E}$ using the attention weights.
This process can be defined by:
\begin{equation}
\begin{aligned}
    \mathbf{\Tilde{h}}_i=\sum_{\mathbf{h}_j \in \mathbf{E}}a_{ij} \mathbf{h}_j, \text{ where } a_{ij} = \frac{\exp(\mathbf{\hat{h}}_i\cdot \mathbf{h}_j)}{\sum_{\mathbf{h}_k\in \mathbf{E}}\exp(\mathbf{\hat{h}}_i\cdot \mathbf{h}_k)}.
\end{aligned}
\end{equation}
where $\mathbf{h}_j$ is the $j$-th embedding from $\mathbf{E}$ and $a_{ij}$ is the attention weight between $\mathbf{\hat{h}}_i$ to $\mathbf{h}_j$.
Later, we concatenate $\mathbf{\hat{h}}_i$  and $\mathbf{\Tilde{h}}_i$ together and input them into a fully-connected layer activated by Softmax to estimate the distribution of each feature ID token for the $i$-th step, which can be formulated by:
\begin{equation}
    P_{\psi}(f_i|\mathbf{E}, \mathbf{f}_{<i}) = \frac{\exp(\mathbf{W}_{\mathbf{h}_i}\text{ concat}(\mathbf{\tilde{h}}_i,\mathbf{\hat{h}}_i))}{\sum_{\mathbf{h}_k \in \mathbf{E}}\exp({\mathbf{W}_{\mathbf{h}_k}\text{ concat}(\mathbf{\tilde{h}}_i,\mathbf{\hat{h}}_i))}},
\end{equation}
where $\text{concat}(\cdot)$ is the concatenate operation, $\mathbf{W}_{(\cdot)}$ represents the weight matrix. $\mathbf{f}_{< i}$ represent the previous tokens before the $i$-th step.
We may take the token with the highest probability as the predicted feature ID token, denoted by $f_i$.
 If multiple the probability at each step, the probability distribution of the whole token sequence $\mathbf{f}$ can be represented by:
\begin{equation}
P_{\psi}(\mathbf{f}|\mathbf{E}) = \prod_{t=1}^{T}P_{\psi}(f_t|\mathbf{E}, \mathbf{f}_{<t}).
\end{equation}

\smallskip
\noindent\textbf{The Feature Subset Evaluator $\omega$:} 
The evaluator $\omega$ aims to predict the model performance given the continuous embedding of $\mathbf{E}$.
Specifically, we apply the mean pooling to the embedding $\mathbf{E}$ in a column-wise manner to obtain the embedding $\mathbf{\Bar{\mathbf{e}}}\in\mathbb{R}^{1\times d}$.
We then input it into a fully-connected layer to predict the model performance $\Ddot{v}$, which can be defined by:
$
    \Ddot{v} = \omega(\bar{\mathbf{e}}).
$

\smallskip
\noindent\textbf{The joint optimization.}
We jointly optimize the encoder, decoder, and evaluator.
There are two goals:
1)  reconstructing the feature ID tokens $\mathbf{f}$.
To achieve this goal, we minimize the negative log-likelihood of $\mathbf{f}$ given $\mathbf{E}$, defined by:
\begin{equation}
\mathcal{L}_{rec}=-\log P_{\psi}(\mathbf{f}|\mathbf{E})=-\sum_{t=1}^{T} \log P_{\psi}(f_t|\mathbf{E}, \mathbf{f}_{<t}).
\end{equation}
2) minimizing the difference between the predicted  performance $\Ddot{v}$ and the real one $v$.
To achieve this goal, we minimize the 
 Mean Squared Error (MSE) between $\Ddot{v}$ and $v$, defined by:
\begin{equation}
    \mathcal{L}_{est} = \text{MSE}(v, \Ddot{v}).
    \label{p_loss}
\end{equation}
We integrate the two losses to form a joint training loss $\mathcal{L}$, defined by:
\begin{equation}
        \mathcal{L} = \lambda \mathcal{L}_{rec} + (1-\lambda)\mathcal{L}_{est},
\end{equation}
where $\lambda$ is the trade-off hyperparameter to balance the contribution of the two objectives during the learning process. 

\subsection{Gradient-Optimized Best Embedding Search}
\noindent\textbf{Why selecting search seeds matters.}
When the encoder-evaluator-decoder model converges, we perform a gradient-based search in the well-built continuous space for better feature selection results.
Similar to the significance of initialization for deep neural networks, good starting points can accelerate the search process and enhance its performance.
These points are called ``search seeds''.
Thus, we first rank the selection records based on the corresponding model performance.
Then, the top-K selection records are  chosen as search seeds for searching for better embeddings. 

\smallskip
\noindent\textbf{Gradient-ascent Optimizer Search.}
Assuming that one of the top-K selection records is 
$(\mathbf{f}, v)$.
We input $\mathbf{f}$ into the encoder to obtain the embedding $\mathbf{E}=[\mathbf{h}_1, \mathbf{h}_2,\cdots, \mathbf{h}_T]$.  
Then, we move each embedding in $\mathbf{E}$ along the gradient direction induced by the evaluator $\omega$:
\begin{equation}
    \mathbf{h}_t^+=\mathbf{h}_t + \eta \frac{\partial \omega}{\partial \mathbf{h}_t}, \mathbf{E}^+=\{\mathbf{h}_1^+,\mathbf{h}_2^+,\cdots,\mathbf{h}_T^+,\},
\end{equation}
where $\eta$ is the step size and $\mathbf{E}^+$ is the enhanced embedding.
We should set $\eta$ within a reasonable range (e.g., small enough) to make the model performance of the enhanced embedding $\mathbf{E}^+$ is better than $\mathbf{E}$, denoted by $\omega(\mathbf{E}^+) \geq \omega(\mathbf{E})$.
We conduct the search process for each record and collect these enhanced embeddings, denoted by $[\mathbf{E}^+_1,\mathbf{E}^+_2,\cdots,\mathbf{E}^+_K]$.
Since we have embedded discrete selection records into $d$ dimensional embedding space, the time complexity of the searching process is independent of the size of the original feature set.

\subsection{Optimal Feature Subset Reconstruction}
The enhanced embeddings indicate possible better feature selection results.
To find the optimal one, we should reconstruct the feature ID tokens using them.
Specifically, we input $[\mathbf{E}^+_1,\mathbf{E}^+_2,\cdots,\mathbf{E}^+_K]$ into the well-trained decoder to get the reconstructed feature ID tokens $[\mathbf{f}_1^+,\mathbf{f}_2^+,\cdots,\mathbf{f}_K^+]$.
We do not need to set the number of feature ID tokens throughout the decoding process; instead, we identify each token until the stop token, similar to how natural language generation works.
Then, we use them to select different feature subsets from the original feature set. 
Finally, we adopt the downstream ML model to  evaluate these subsets and output the optimal feature ID tokens $\mathbf{f}^*$, which can be used to select the best feature subset with the highest model performance.

\section{Experiments}
In this section, we present detailed experimental setups and conduct comprehensive experimental analyses and case studies to validate the efficacy of the proposed model.

\subsection{Experimental Setup}

\noindent{\bf Dataset Description.}
We conducted extensive experiments using 19 publicly available datasets from UCI, OpenML, CAVE, Kaggle, and LibSVM.
These datasets are categorized into 3 folds based on the types of ML tasks: 1) binary classification (C); 2) multi-class classification (MC); 3)  regression (R).
Table \ref{table_overall_perf} shows the statistics of these datasets, which is accessible via the website address provided in the Abstract. 

\begin{table*}[!htbp]
\centering
\caption{Overall performance comparison. `C' for binary classification, `MC' for multi-class classification, and `R' for regression. The best results are highlighted in \textbf{bold}. The second-best results are highlighted in \underline{underline}. (\textbf{Higher values indicate better performance.})}
\label{table_overall_perf}
\resizebox{\linewidth}{!}{
\setlength{\tabcolsep}{0.75mm}{
\begin{tabular}{cccccccccccccccc}
\toprule
Dataset          & Task & \#Samples & \#Features & Original & K-Best  & mRMR & LASSO & RFE  & LASSONet &GFS &SARLFS  & MARLFS &RRA & MCDM & \model        \\  \midrule
SpectF & C   & 267     & 44 &  75.96 &78.21&  79.16&  75.96&  \underline{80.80} &  75.96& 75.01 &75.96&  75.96& 79.16 & 80.36 & \textbf{87.38} \\  
SVMGuide3 & C   & 1243    & 21&  77.81 &75.13&  75.34&  75.95&  78.07&  76.44& \underline{83.12} &  79.48&  81.32  & 77.63 & 76.66 & \textbf{83.68} \\  
German Credit      & C   & 1001    & 24 & 64.88 &  66.67&  65.81&  69.65  &  64.86&   58.20& 67.54 & 63.32&  69.00& 69.69 & \underline{70.85} &  \textbf{75.34} \\ 
Credit Default     & C   & 30000   & 25  & 80.19 &  \underline{80.59}&  \underline{80.59}&  77.94&  80.28&   80.05& 79.96 &  80.05&  80.24&  75.39 & 74.46 & \textbf{80.61}\\ 
SpamBase          & C   & 4601    & 57 & 92.68 & 92.02&  91.91&  91.74&  91.68&   88.39& 92.25 &90.94&  \underline{92.35}&  89.43 & 88.95 & \textbf{92.93}  \\  
Megawatt1  & C & 253 & 38 &  81.60 &78.55&  80.08&  83.78 &  80.08 &  81.03& 77.42 & 82.75&  79.33&  \underline{87.78} & 85.11 &\textbf{90.42}\\
Ionosphere        & C   & 351     & 34 & 92.85 &  91.38&  \underline{95.69}&  86.98&  \underline{95.69} &   85.58& 91.34 &94.27&  88.48&   92.74 & 88.64 &\textbf{97.10}\\  
\midrule
Activity  & MC & 10299 & 561 & 96.17 & 96.07&  95.92&  95.92&  95.87&  \underline{96.17} & 96.12 &95.87&  95.87& 95.58 & 96.12 &  \textbf{96.46}\\
Mice-Protein  & MC & 1080 & 77 & 74.99 & \underline{78.69} &  76.84&  78.71&  77.29&  78.23& 77.35 &74.53&  77.30& 70.86 & \underline{78.69} &  \textbf{79.16} \\
Coil-20   & MC & 1440 & 400 &  96.53 & 95.84 &  95.49&  95.84&  \underline{96.52} & 82.97& 95.49 &   94.79&  95.47& 95.15 & 95.50 &\textbf{97.22}\\
MNIST   & MC & 10000 & 784 &92.70 &\underline{92.75}&  92.35&  92.20&  92.25&  87.05& 91.60 &  91.70&  91.75& 90.40 & 92.65 & \textbf{93.20}\\
MNIST fashion   & MC & 10000 & 784 & 80.15 & \underline{80.80} &  80.00&  79.90&  80.50 &  78.85& 80.60 &80.10&  80.15& 78.85 & 80.20 &  \textbf{81.00}\\
\midrule
Openml\_586         & R   & 1000    & 25 & 54.95 & 57.68&  57.29&  60.67  &  58.10 &  58.60& \underline{63.27} & 55.76&  57.21&  57.80 & 57.95 & \textbf{64.00}  \\
Openml\_589          & R   & 1000    & 25 & 50.95 &54.09&  54.03&  \underline{59.74} &  54.25&  54.80& 44.72 & 38.64&  52.77& 54.54 & 55.43 &   \textbf{59.76} \\  
Openml\_607         & R   & 1000    & 50 & 51.74 & 53.03&  53.03&  \underline{58.10} &  54.39&  53.63& 45.70 & 55.32&  53.94& 55.41 &  55.56 &  \textbf{66.01} \\  
Openml\_616           & R   & 500     & 50 & 15.63 & 24.75&  24.44&  \underline{28.98} &  24.08&  16.32& 52.93 & 25.37&  25.83& 23.32 & 22.92  &  \textbf{47.39} \\  
Openml\_618          & R   & 1000    & 50 & 46.89 & 51.33&  51.33&  47.41&  50.64 &  50.69& \underline{52.40} & 50.18&  51.71 & 50.93 & 50.90 &  \textbf{59.13} \\  
Openml\_620        & R   & 1000    & 25 & 51.01 & 53.57&  53.57&  57.99 &  53.96 &  54.29& \underline{61.99} & 34.45&  53.68& 56.24& 55.66 &  \textbf{62.58} \\  
Openml\_637          & R   & 500     & 50 & 14.95 & 20.72&  20.72&  26.02 &  17.82&  18.97& \underline{40.12} & 19.54&  23.72& 22.38 & 22.16 &  \textbf{42.12} \\  
  \bottomrule
\end{tabular}}}
 \begin{tablenotes}
    \small
    \item * We reported F1-Score for classification tasks, Micro-F1 for multi-class classification tasks, and 1-RAE for regression tasks.
    \end{tablenotes}
\end{table*}
\begin{figure*}[!h]
\centering
\subfigure[Spectf]{
\includegraphics[width=4.25cm]{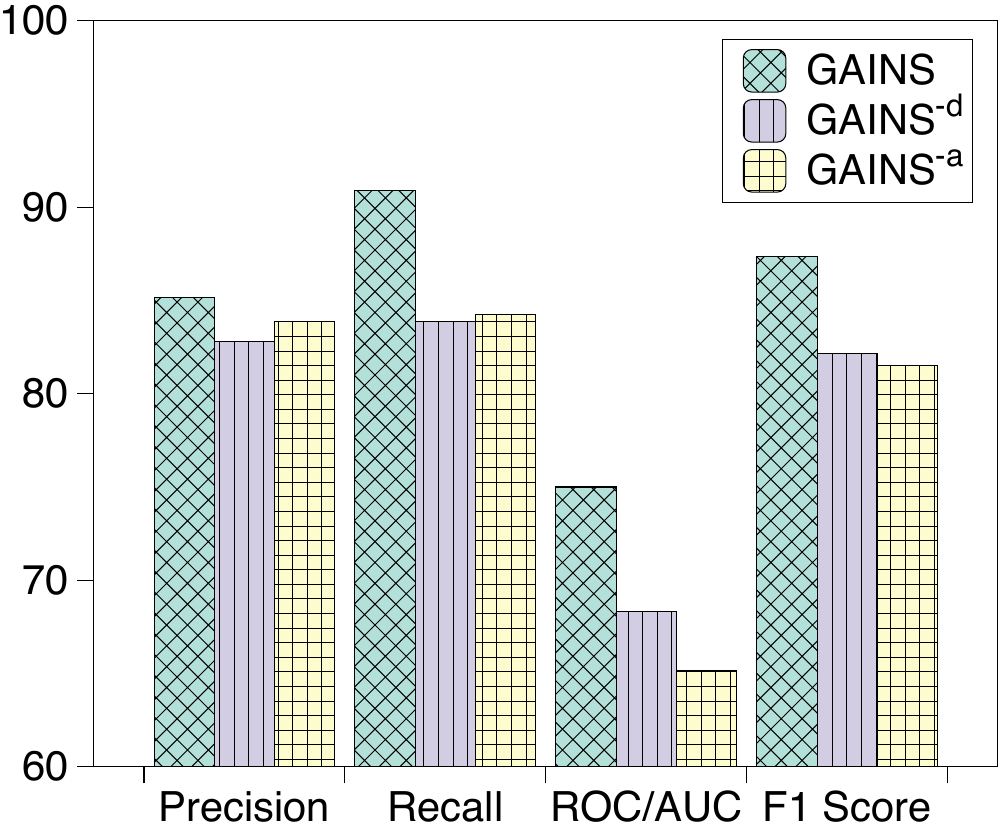}
}
\hspace{-3mm}
\subfigure[German Credit]{
\includegraphics[width=4.25cm]{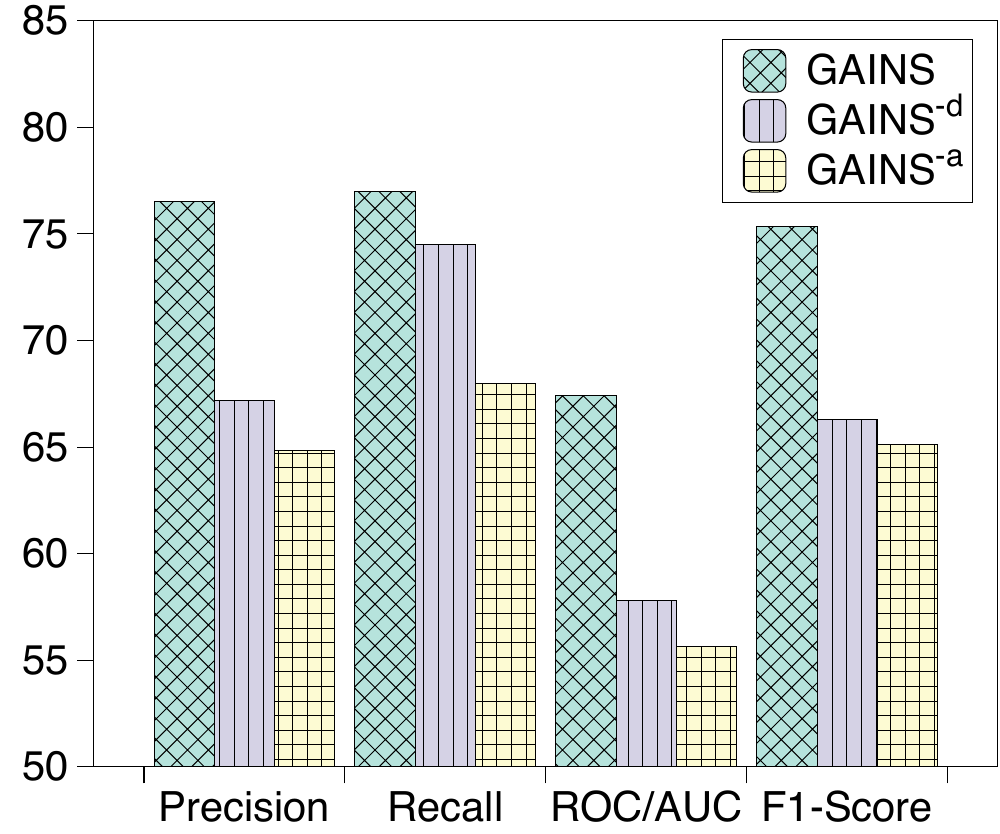}
}
\hspace{-3mm}
\subfigure[OpenML\_589]{ 
\includegraphics[width=4.25cm]{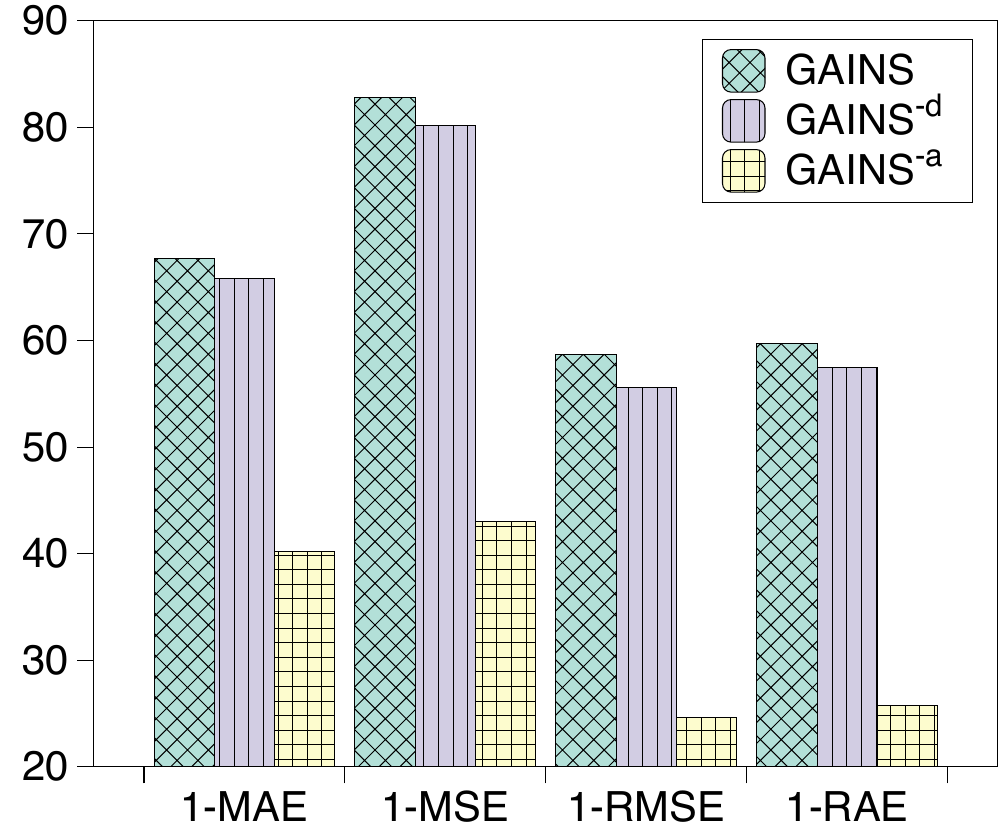}
}
\hspace{-3mm}
\subfigure[Mice Protein]{ 
\includegraphics[width=4.25cm]{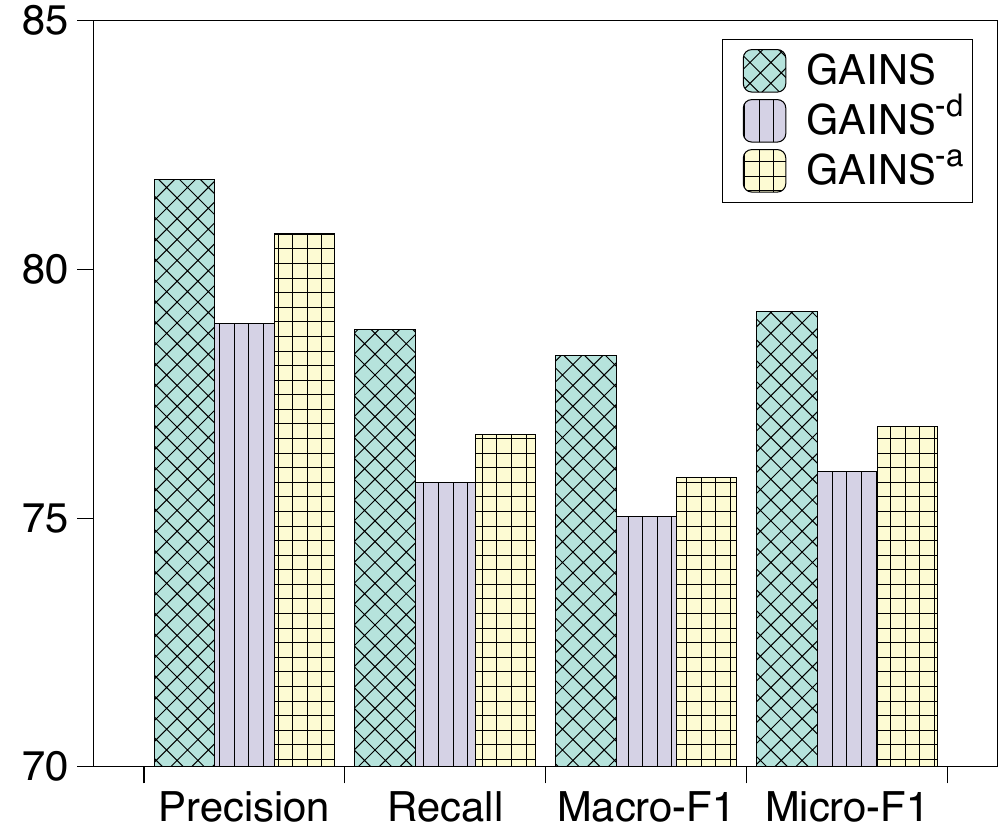}
}
\caption{The influence of data collection (\model$^{-d}$) and data augmentation (\model$^{-a}$) in \model.}
\label{w/oaug}
\end{figure*}

\smallskip
\noindent{\bf Evaluation Metrics.}
For the binary classification task, we adopted F1-score, Precision,  Recall, and  ROC/AUC.
For the multi-classification task, we used Micro-F1, Precision, Recall, and Macro-F1.
For the regression task, we utilized 1-Mean Average Error (1-MAE),  1-Mean Square Error (1-MSE), and  1-Root Mean Square Error (1-RMSE).
For all metrics, the higher the value is, the better the model performance is. 


\smallskip
\noindent{\bf Baseline Algorithms.}
We compared {\model} with 10 widely used feature selection methods: (1) \textbf{K-Best} selects K features with the highest feature scores~\cite{kbest}; (2) \textbf{mRMR} intends to select a feature subset with the greatest relevance to the target and the least redundancy among themselves~\cite{mrmr}; (3) \textbf{LASSO} selects features by regularizing model parameters, which shrinks the  coefficients of useless features into zero~\cite{lasso}; (4) \textbf{RFE} recursively removes the weakest features until the specified number of features is reached~\cite{rfe}; 
(5) \textbf{LASSONet} designs a novel objective function to conduct feature selection in neural networks~\cite{lassonet}; (6) \textbf{GFS} selects features using genetic algorithms, which recursively  
generates a population based on a possible feature subset, then uses a predictive model to evaluate it ~\cite{gfe};
(7) \textbf{MARLFS} builds a multi-agent system for selecting features, wherein each agent is associated with a single feature, and feature redundancy and downstream task performance are viewed as incentives~\cite{marlfs};
(8) \textbf{SARLFS} is a simplified version of MARLFS, which uses  one agent to determine the selection of all features in order to alleviate computational costs~\cite{sarlfs};  
(9) \textbf{RRA} first collects distinct selected feature subsets, and then integrates them based on statistical sorting distributions ~\cite{seijo2017ensemble}; 
(10) \textbf{MCDM} first obtains a decision matrix using the ranks of features, then assigns feature scores based on the matrix for ensemble feature selection~\cite{mcdm}.
Among the discussed baseline models, K-Best and mRMR fall under the category of filter methods. Lasso, RFE, and LassoNet are considered as embedded methods. GFS, SARLFS, and MARLFS are classified as wrapper methods. Lastly, RRA and MCDM are representative of hybrid feature selection methods.

We randomly split each dataset into two independent sets. The prior 80\% is used to build the embedding space, and the remaining 20\% is used to search for the optimal feature space.
We conducted all experiments using the hold-out setting to ensure a fair comparison. 
We adopted Random Forest as the downstream machine learning model and reported the performance of each method by running five-fold cross-validation on the testing set. 
Random Forest is a robust, stable, well-tested method, thus, we can reduce performance variation caused by the model, and make it easy to study the impact of the result of feature selection.  

\noindent{\bf Hyperparameter Settings and Reproducibility.}
We ran MARLFS for 300 epochs to collect historical feature subsets and the corresponding downstream task performance.
To augment data, we permutated the index of each feature subset 25 times.
These augmented data can be used for {\model } training.
The Encoder and Generator have the same model structure, which is a single-layer LSTM.
The Predictor is made up of 2-layer feed-forward networks.
The hidden state sizes of the Encoder, Decoder, and Predictor are 64, 64, and 200, respectively.
The embedding size of each feature index is 32.
To train {\model }, we set the batch size as 1024, the learning rate as 0.001, and $\lambda$ as 0.8 respectively. 
During the model inference stage, we used the top 25 
feature selection records as the initial points to search for the optimal feature space. 

\noindent{\bf Environmental Settings}
All experiments were conducted on the Ubuntu 18.04.6 LTS operating system, AMD EPYC 7742 CPU, and 8 NVIDIA A100 GPUs, with the framework of Python 3.9.10 and PyTorch 1.8.1~\cite{paszke2019pytorch}.

\begin{figure*}[!t]
\centering
\hspace{-4mm}
\subfigure[Parameter Size (MB)]{
\includegraphics[width=4.3cm]{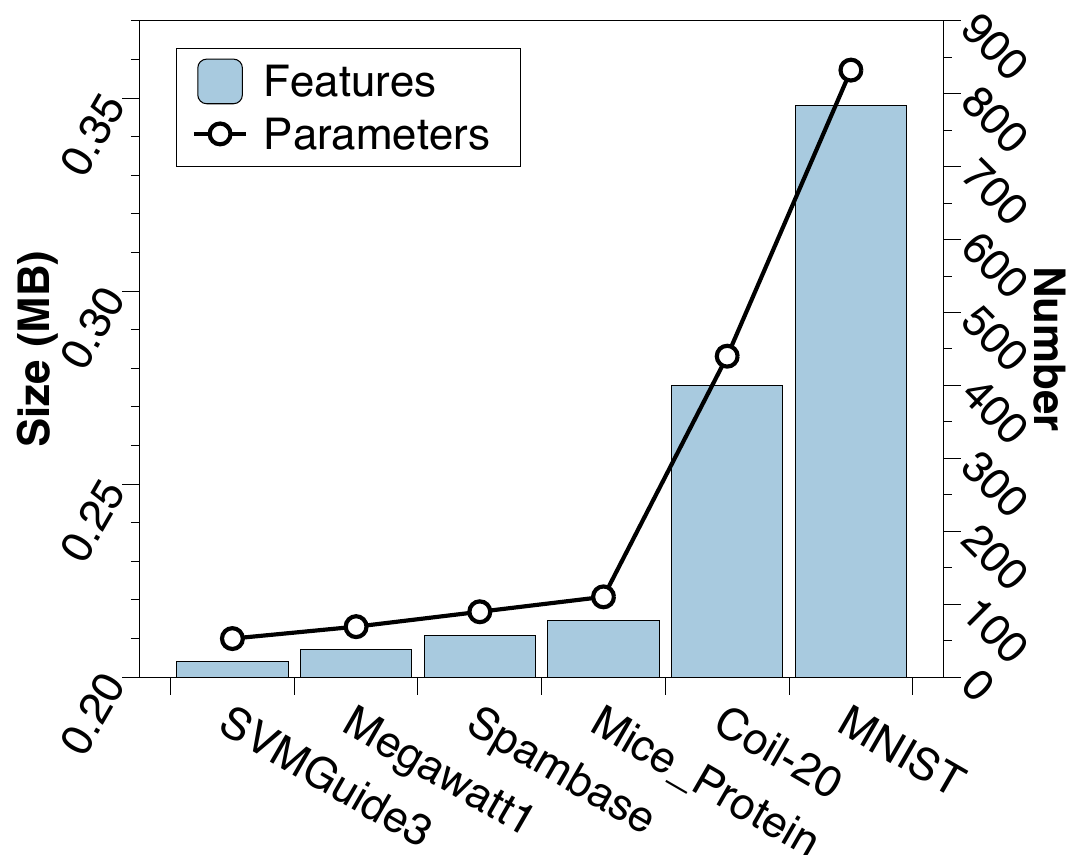}
}
\hspace{-3mm}
\subfigure[Inference Time (s)]{
\includegraphics[width=4.3cm]{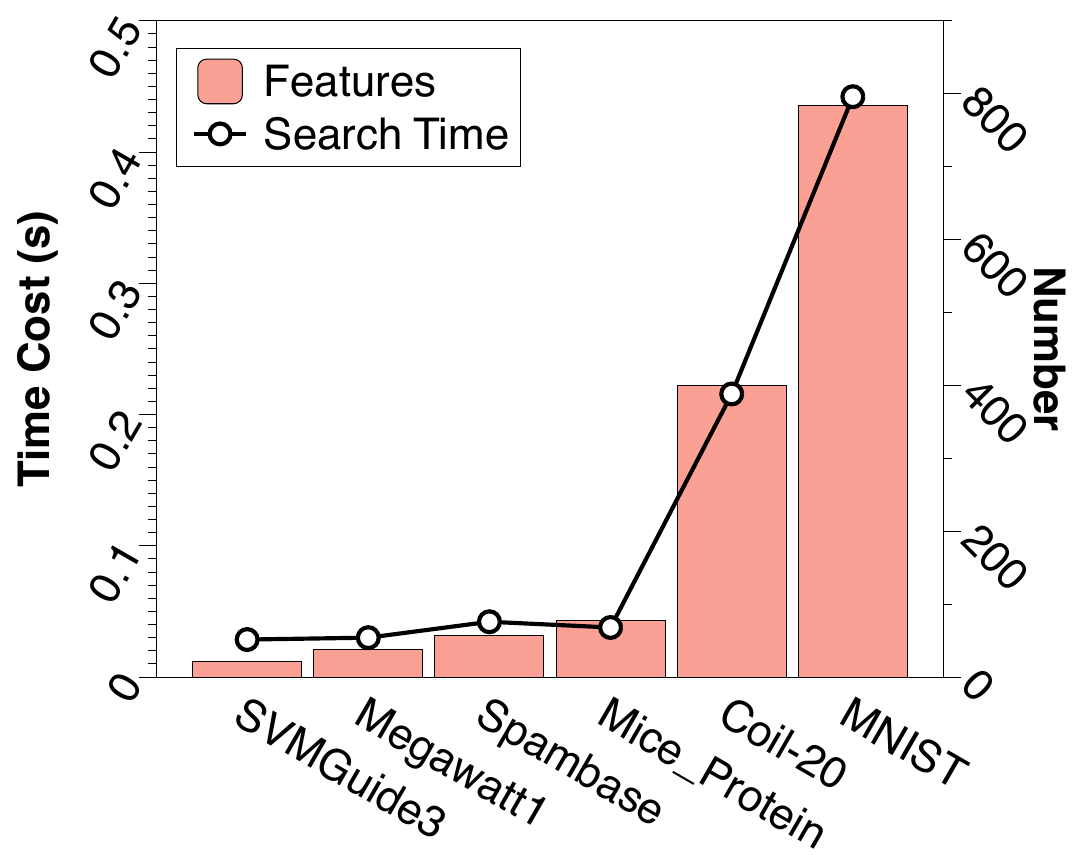}
}
\hspace{-3mm}
\subfigure[Data Preparation Time (s)]{
\includegraphics[width=4.3cm]{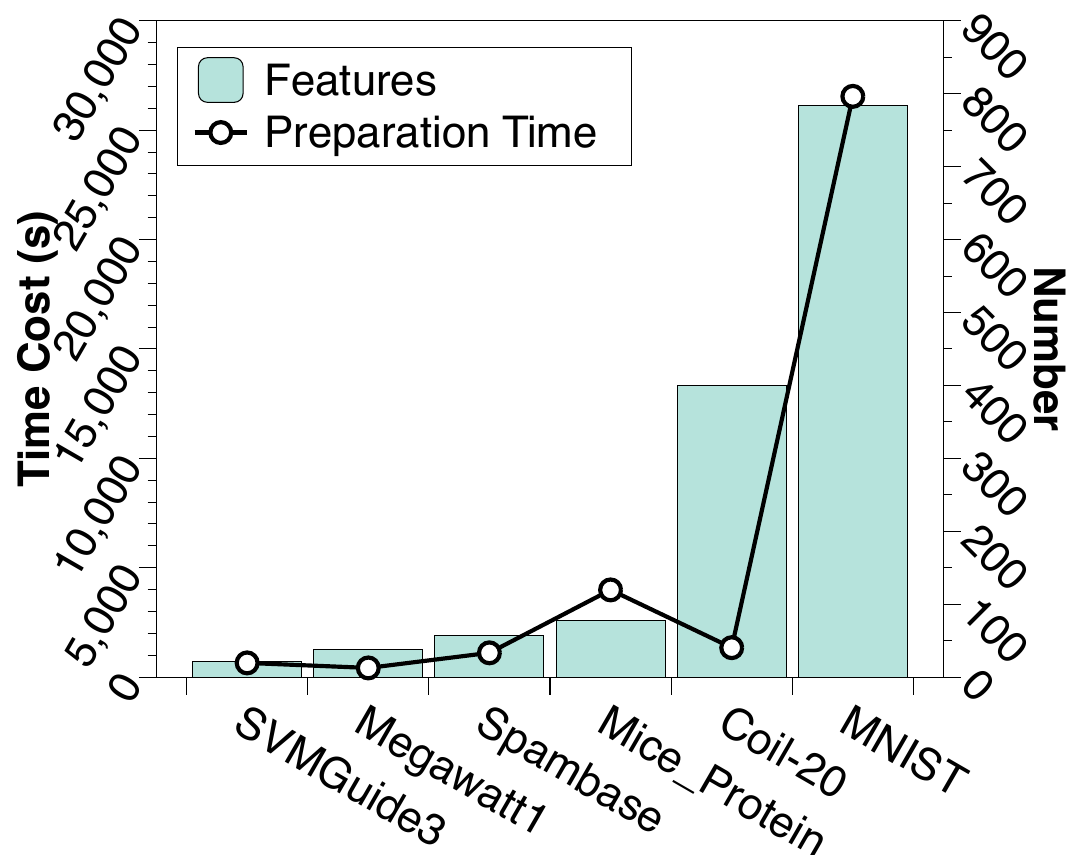}
}
\hspace{-3mm}
\subfigure[Training Time (s)]{
\includegraphics[width=4.3cm]{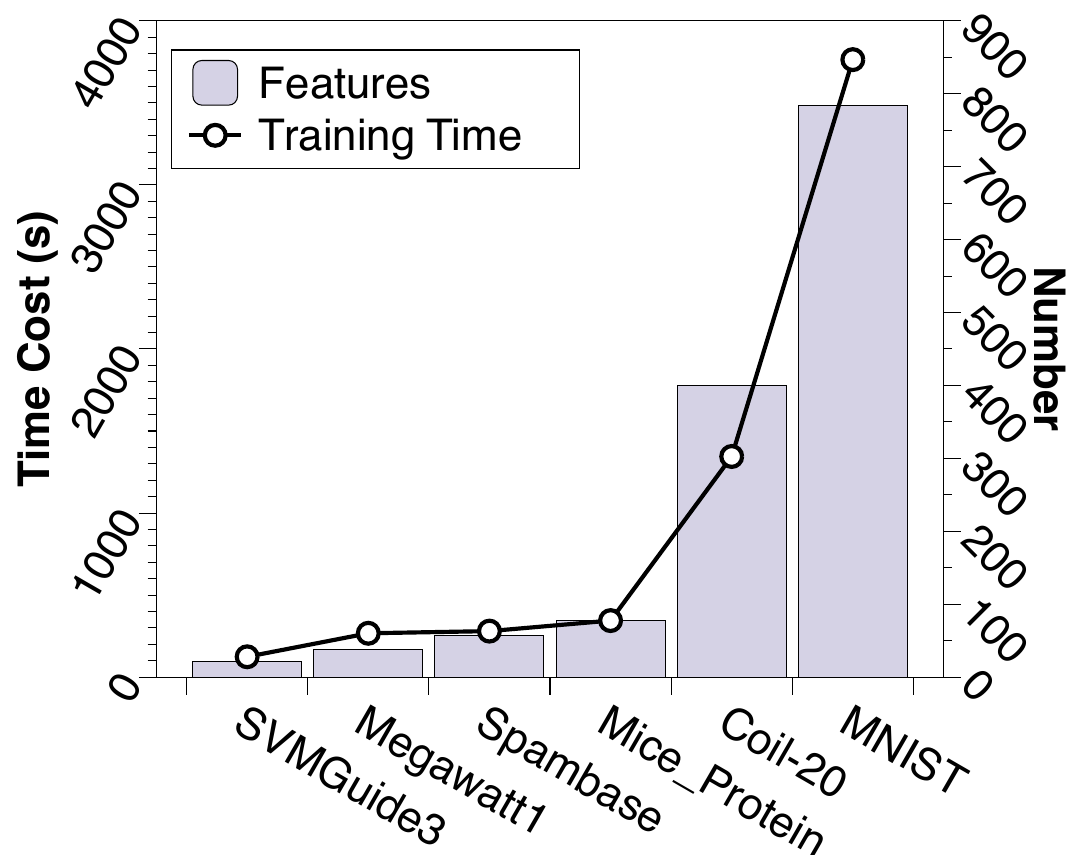}
}
\hspace{-4mm}
\caption{Scalability check of \model\  in terms of feature size, parameter size, inference time, data preparation time,  and training time.}
\label{scalable}
\end{figure*}

\subsection{Experimental Results}

\noindent{\bf Overall Comparison.}
This experiment aims to answer: \textit{Can \model\ effectively select the feature subset with excellent performance?}
Table~\ref{table_overall_perf} shows the overall comparison results in terms of F1-score, Micro-F1, and 1-RAE. 
We observed that \model\ outperforms other baseline models across all domains and various tasks.
The underlying driver for this observation is that \model\ converts the discrete feature selection records into a discriminative and effective embedding space by integrating the knowledge of historical selection records.
It enables the gradient-based search to effectively perceive the properties of the original feature set to obtain superior feature subsets.
Another interesting observation is that the performances of various baseline models vary over  datasets: high in certain datasets, yet low in other datasets. 
Such an observation indicates that classic feature selection methods can address the feature selection task in a limited number of data environments, but perform unstably in diverse data environments.
Overall, this experiment demonstrates that \model\ has  effective and robust performance in various data environments and application scenarios.

\smallskip
\noindent{\bf Examining the impact of data collection and augmentation.}
This experiment aims to answer: \textit{Is it essential to collect feature selection records and augment them to maintain \model\ performance?}
To establish the control group, we developed two model variants: 1) \model$^{-d}$, we collected feature selection records at random rather than building basic feature selectors.
2) \model$^{-a}$, we removed the data augmentation process of \model.
Figure~\ref{w/oaug} shows the comparison results on Spectf, German Credit, OpenML\_589, and Mice Protein.
We found that the performance of \model\ is much better than \model$^{-d}$.
The underlying driver is that, compared to random generation, selection records generated by feature selectors are more robust and denoising, which is important for creating a more effective embedding space for searching better feature subsets.
Moreover, we observed that \model\ is superior to \model$^{-a}$ in all cases.
The underlying driver is that data augmentation can increase the data diversity, resulting in a more robust and effective learning process for \model.
Thus, this experiment validates that data collection and 
augmentation is significant for maintaining \model\ performance.

\smallskip
\noindent{\bf Study of the scalability of \model.}
This experiment aims to answer: \textit{Does \model\ have excellent scalability in different datasets?}
According to the dimensionality of the feature set, we chose 6 datasets: SVMGuide3, Megawatt1, Spambase, Mice Protein, Coil-20, and MNIST, from small to large.
We analyzed the variation in  the average time required to make an inference during a single search step and in the parameter size employed by the encoder-evaluator-decoder model.
Figure~\ref{scalable} (a-b) shows the comparison results in terms of Parameter Size and Inference Time.
We found that in spite of the almost 40-fold increase in feature dimension from SVMGuide3 to MNIST, the inference time increases by just about 4-fold, and the parameter size only enlarges by almost 2-fold.
The underlying driver for this observation is that the deep feature subset embedding module has integrated the knowledge of discrete selection records into a fixed continuous embedding space, significantly reducing the searching time and embedding model size. 
The time-cost bottleneck is only caused by the downstream ML model validation. 
Therefore, this experiment validates that \model\ has strong scalability when dealing with datasets of varying dimensions.

\begin{figure*}[!h]
\centering
\subfigure[Spectf]{
\includegraphics[width=4.25cm]{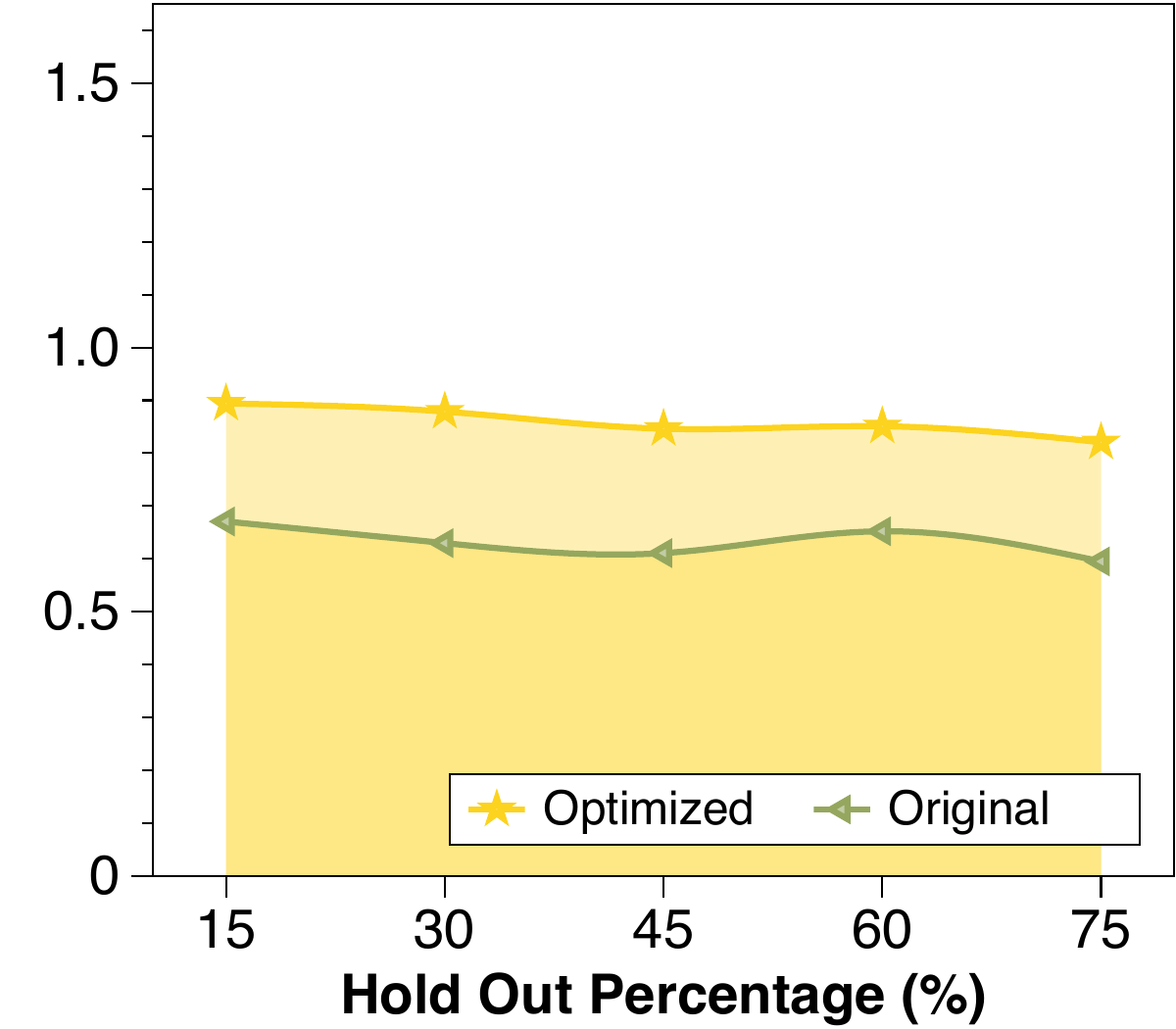}
}
\hspace{-3mm}
\subfigure[Ionosphere]{
\includegraphics[width=4.25cm]{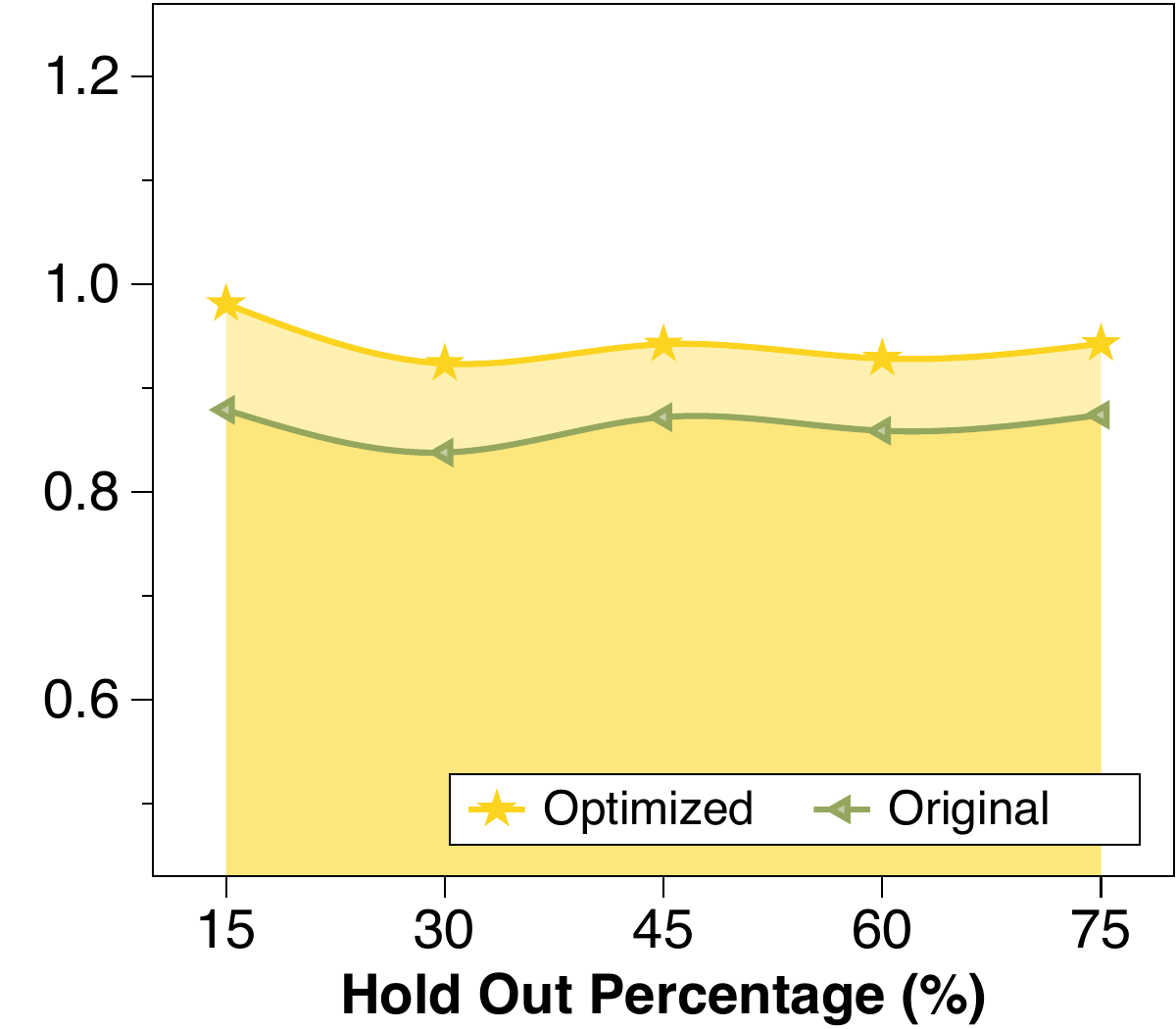}
}
\hspace{-3mm}
\subfigure[OpenML\_607]{ 
\includegraphics[width=4.25cm]{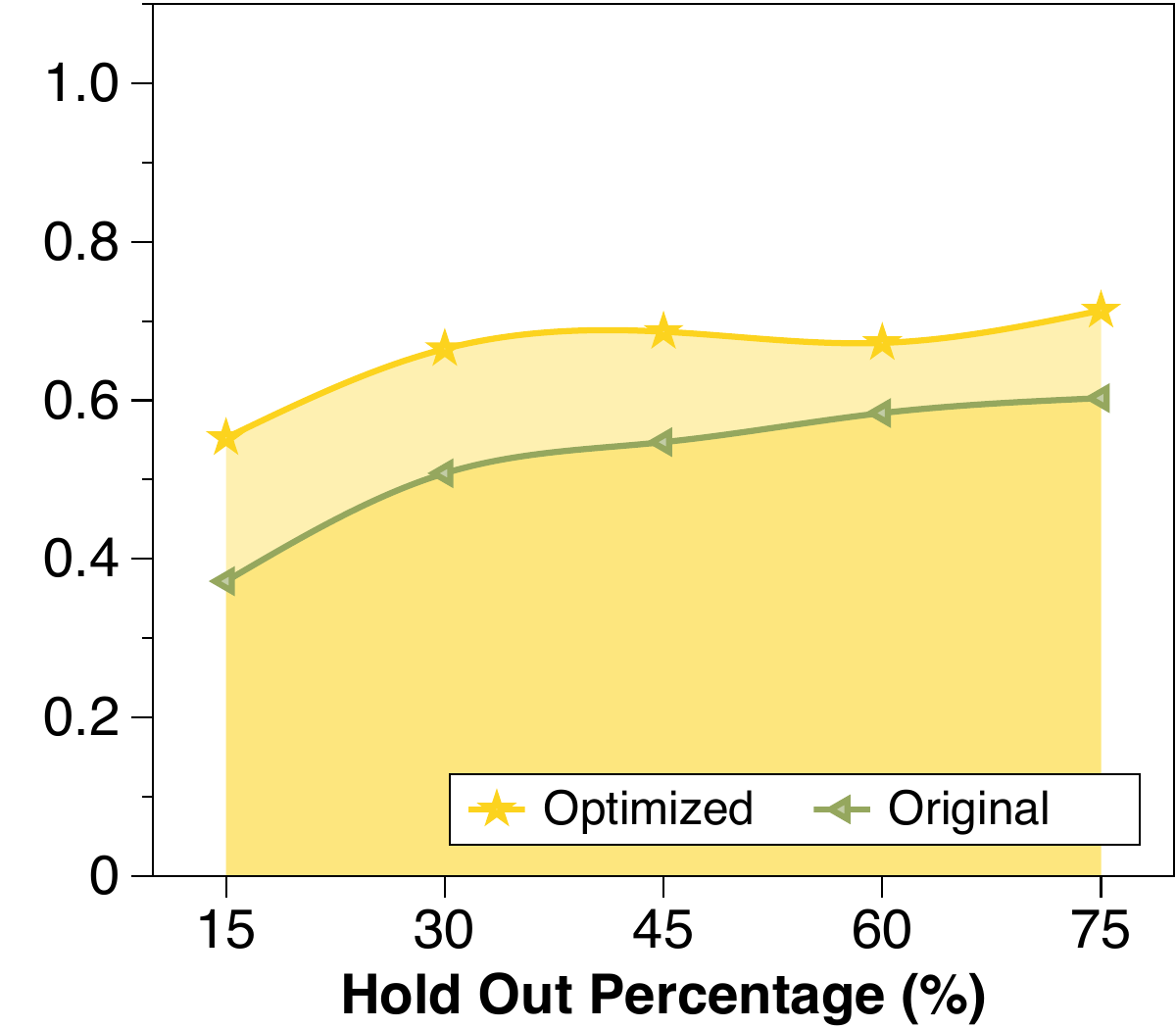}
}
\hspace{-3mm}
\subfigure[Mice Protein]{ 
\includegraphics[width=4.25cm]{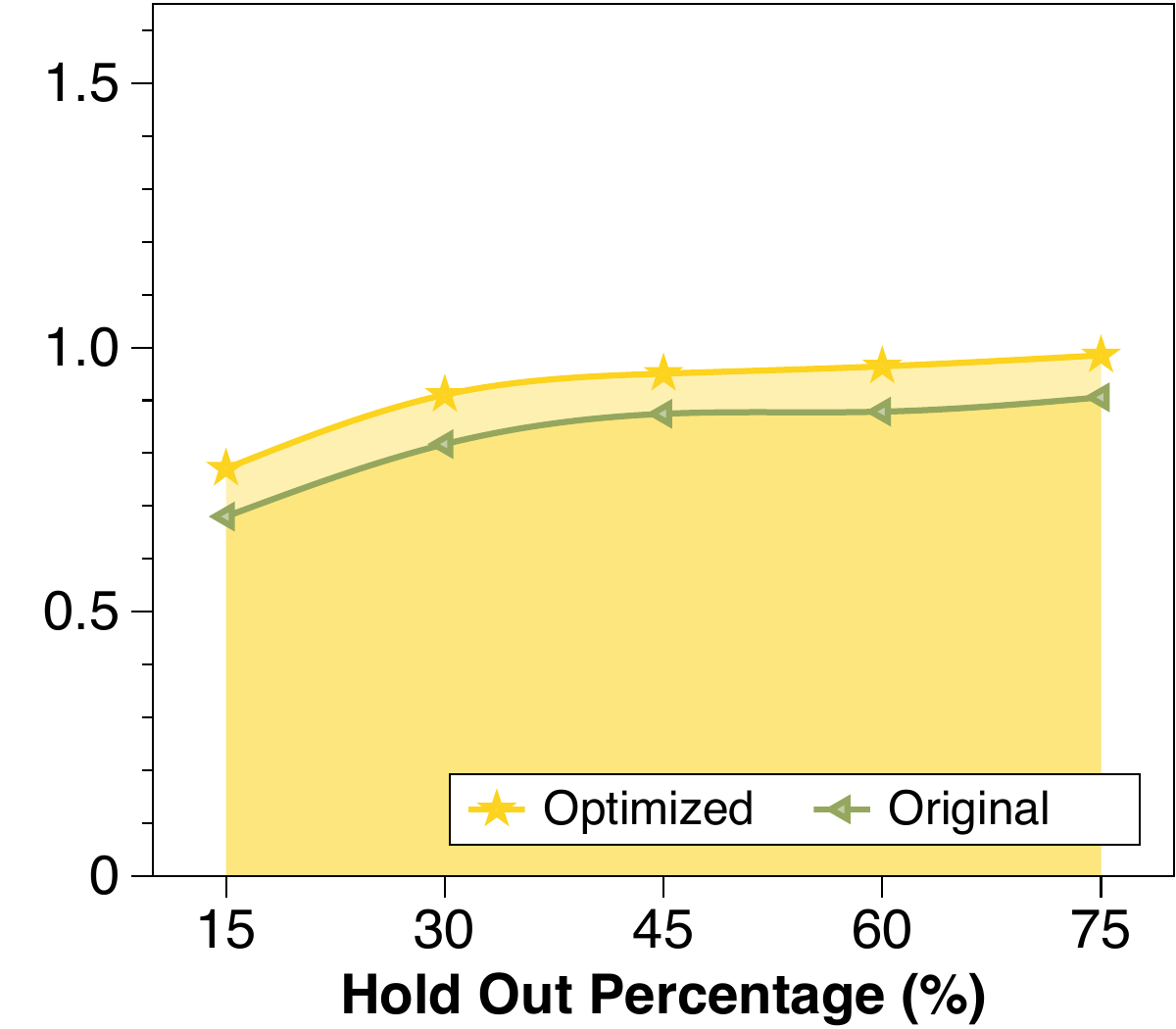}
}
\caption{The influence of variant hold-out percentages in terms of F1-score.}
\label{split}
\end{figure*}

\begin{figure*}[!h]
\centering
\hspace{-3mm}
\subfigure[RandomForest]{
\includegraphics[width=4.25cm]{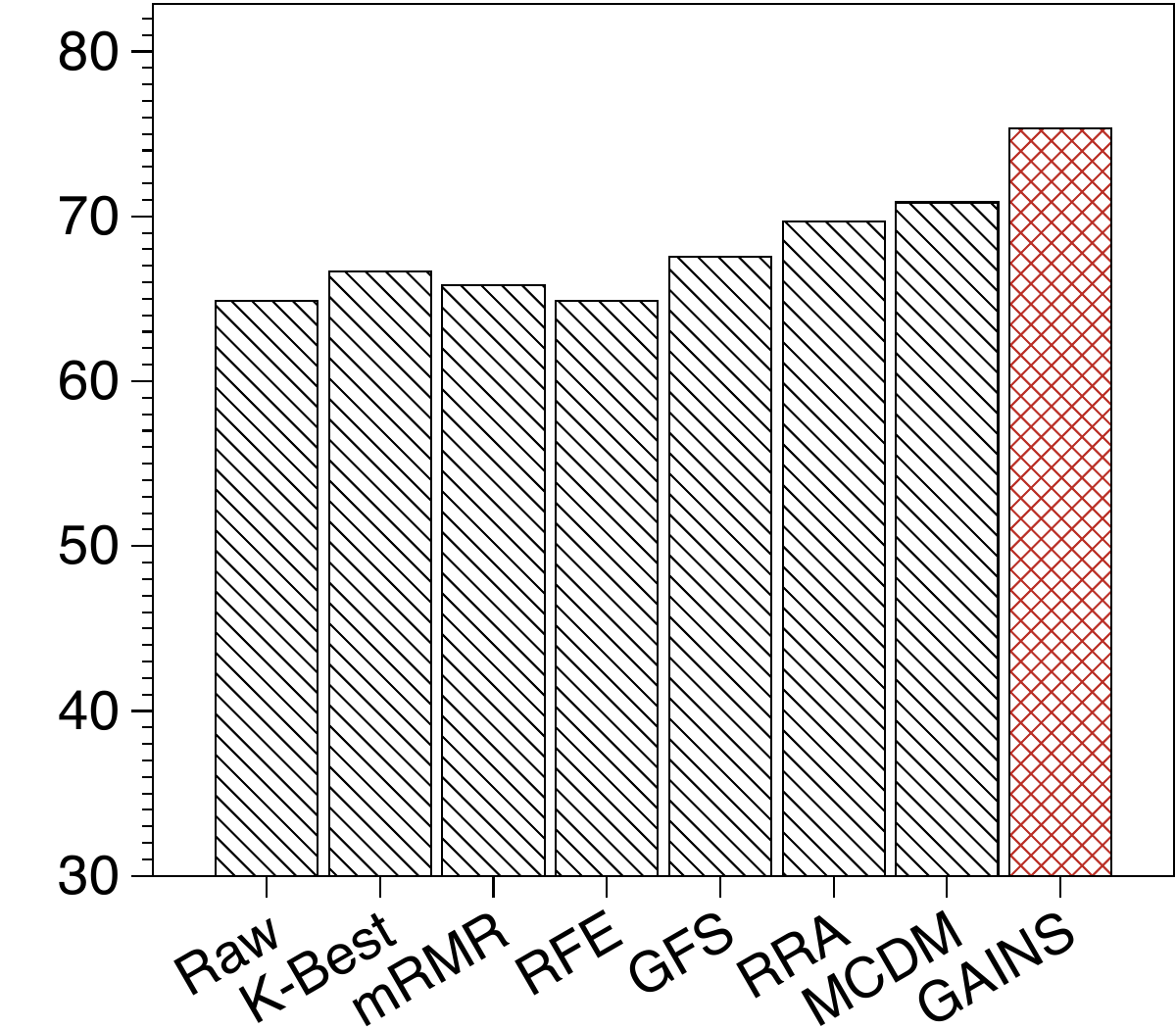}
}
\hspace{-3mm}
\subfigure[XGBoost]{ 
\includegraphics[width=4.25cm]{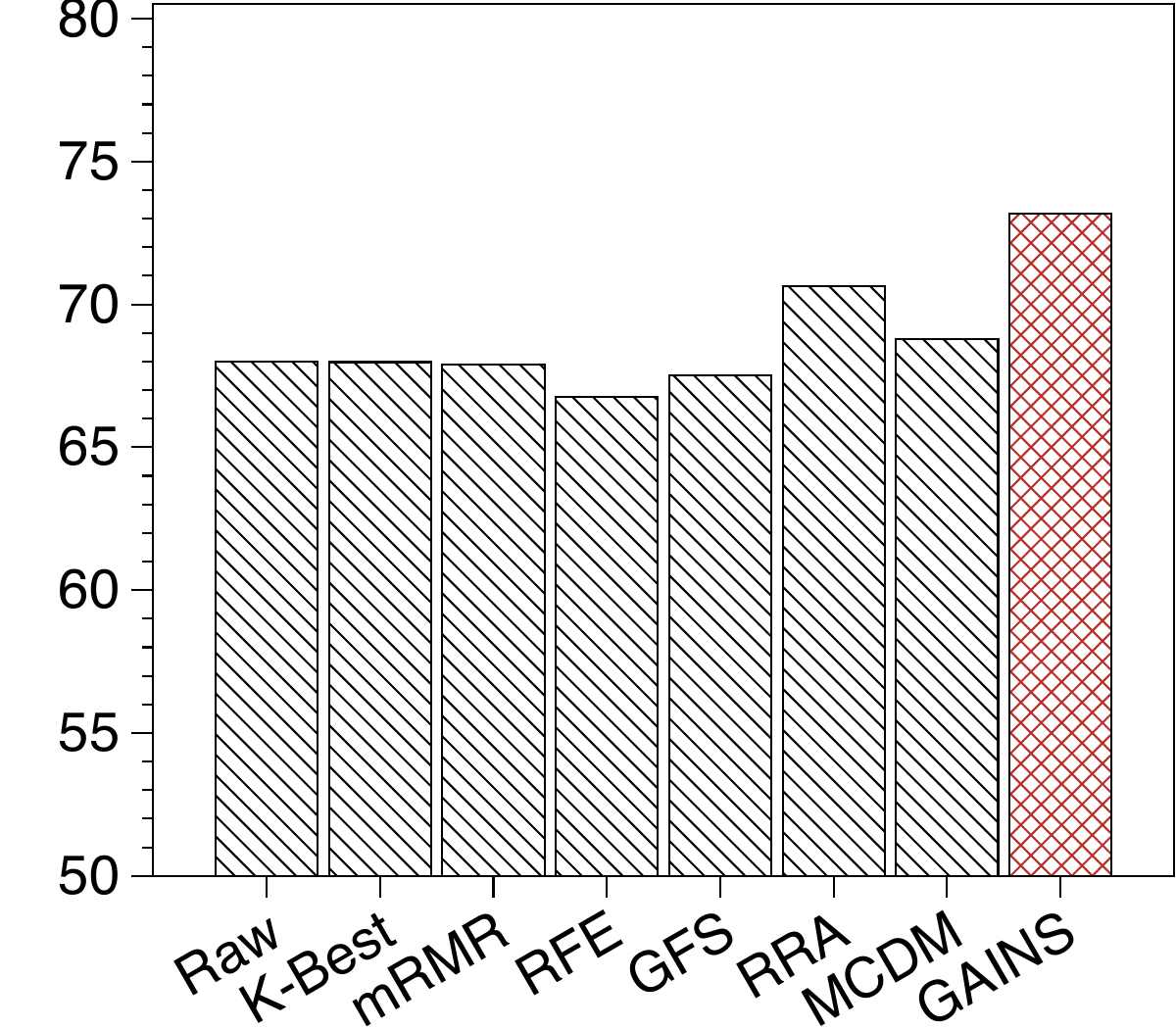}
}
\hspace{-3mm}
\subfigure[SVM]{ 
\includegraphics[width=4.25cm]{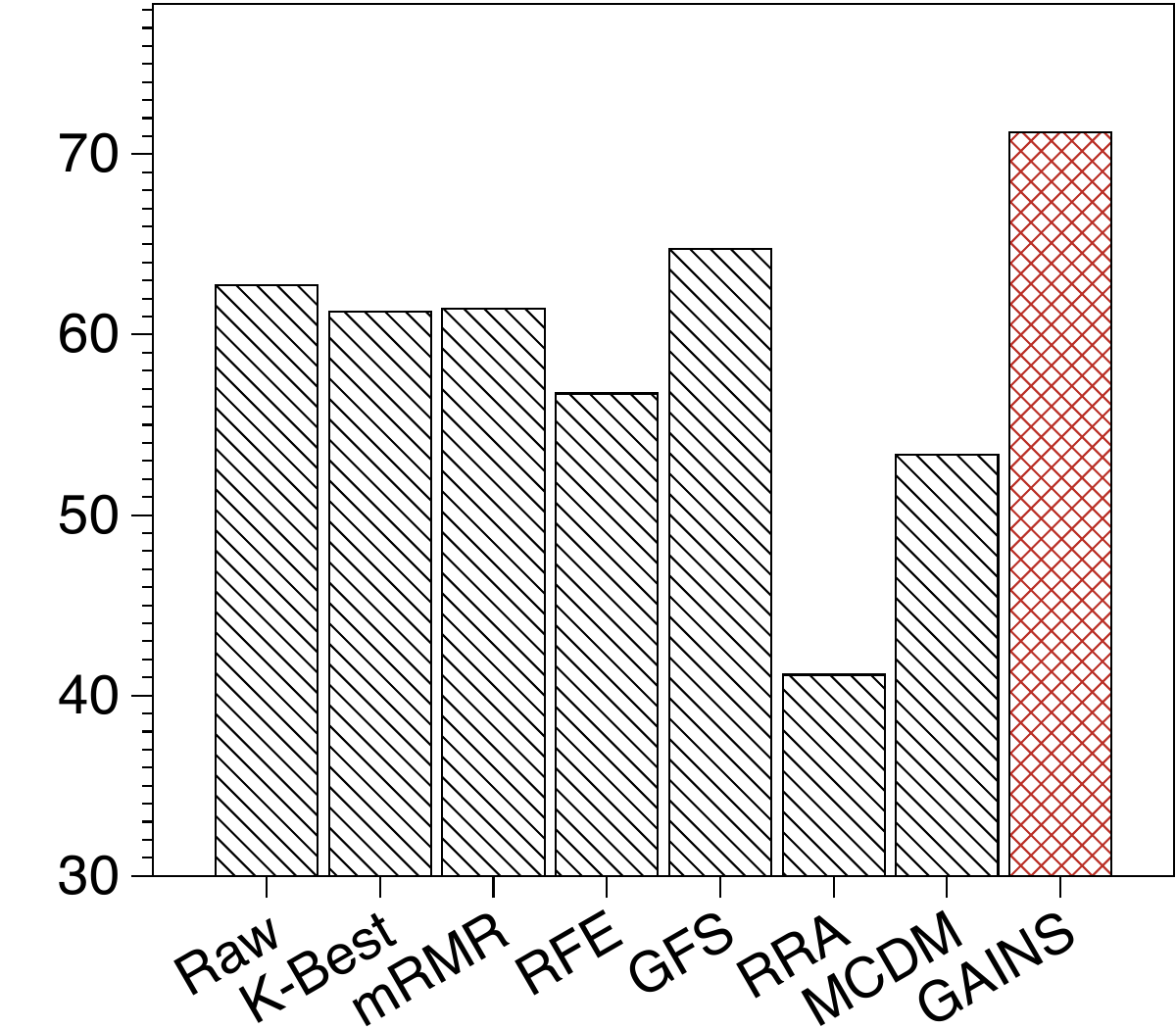}
}
\subfigure[KNN]{
\includegraphics[width=4.25cm]{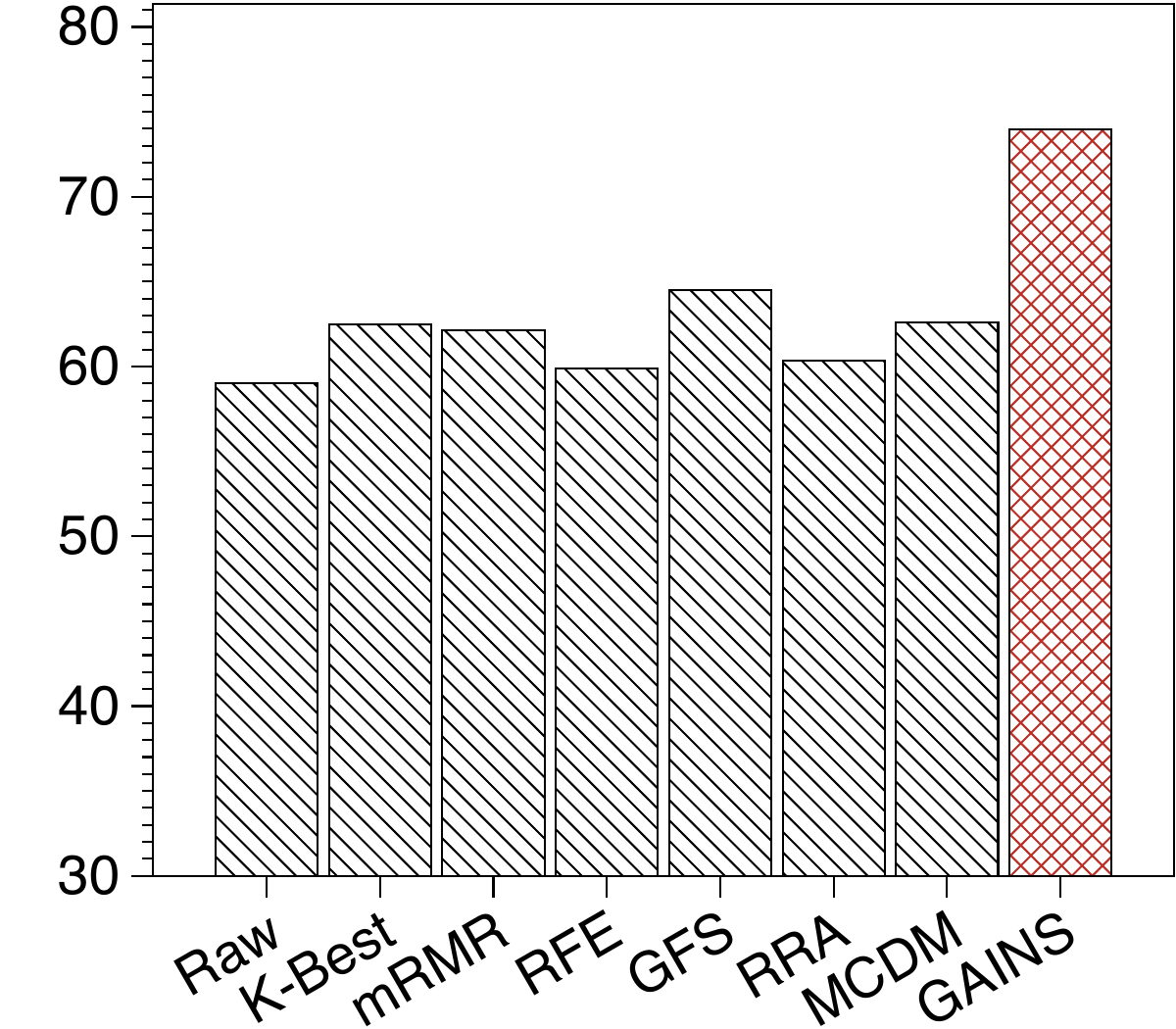}
}
\caption{Robustness check of \model\ when confronted with distinct downstream ML models in terms of F1-score on German Credit.}
\label{table_robust}
\end{figure*}

\smallskip

\noindent{\bf Study of the time cost of data preparation and model training.}
This experiment aims to address the following question: \textit{What are the computational costs in terms of data preparation and model training?} The same as  the scalability examination, we selected six datasets with varying dimensions, i.e., SVMGuide3, Megawatt1, Spambase, Mice Protein, Coil-20, and MNIST. 
Figure~\ref{scalable} (c-d) illustrates the comparison of the corresponding time costs.
From Figure~\ref{scalable} (c) We found that the time cost for data preparation correlates with the feature dimension.
For instance, the MNIST dataset, encompassing 784 feature columns, necessitates approximately 26,547.5 seconds for collecting sufficient samples. In contrast, the Coil-20 dataset, comprising 400 feature columns, demands a lesser duration of 1344.56 seconds. Interestingly, the Mice\_Protein dataset, despite only encompassing 77 feature columns, necessitates 3985 seconds. This observation indicates that the time cost of the reinforcement learning-based data preparation module is associated with the number of feature columns, the sample size, and the complexity of the downstream task.
Figure~\ref{scalable} (d) shows the model training duration increases proportionately with the number of feature columns, thereby validating the findings of the scalability examination. This illustrates that \model\ requires only a finite number of epochs for reinforced data collection, thereby significantly reducing the time cost for exploration relative to RL-based methods such as SARLFS and MARLFS. Secondly, the training time remains unaffected by the sample size or the complexity of the downstream task, indicating that our continuous optimization method can make the feature selection process more efficient.

\smallskip
\noindent{\bf Study of the impact of hold-out percentage.}
This experiment aims to answer: \textit{How does the variation in hold-out settings impact our proposed model?} 
In the experiment setup section,
we partitioned the dataset randomly into two independent subsets. The larger subset, comprising 80\% of the total, was allocated for data preparation, model training, and validation, whereas the remaining 20\% functioned as a holdout set to evaluate the performance of the optimally constructed feature subset. We further explored this holdout setting and conducted an experiment to exemplify its effectiveness, with the results being depicted in Figure~\ref{split}.
Figure~\ref{split} reveals that different datasets display distinct trends. For instance, the Ionosphere dataset's performance is enhanced with a smaller test split. Conversely, the performance on the Spectf dataset remains relatively constant, irrespective of fluctuations in the split setting. Moreover, a smaller test split shows a negative influence on the performance of the OpenML\_607 and Mice\_Protein datasets. Various elements, such as the randomness of the dataset partitioning and the volume of the training data, can affect these datasets' performance.
Notwithstanding, we observed consistently significant improvements between the performance lines of the optimized and original feature sets. This observation indicates that the holdout percentage setting does not influence our proposed method's optimization for the original datasets, thus substantiating the model's robustness.
Therefore, his experiment demonstrates that \model\ can comprehend the feature selection knowledge and produce better feature selection results regardless of the hold-out percentages.

\smallskip
\noindent{\bf Robustness check of {\model } over downstream MLtasks.}
This experiment aims to answer: \textit{Does our proposed model exhibit robustness when confronted  with various machine learning models serving as downstream tasks?} We proceeded to substitute the downstream machine learning model with Random Forest (RF), XGBoost (XGB), Support Vector Machine (SVM), and K-Nearest Neighborhood (KNN) respectively. 
Figure~\ref{table_robust} shows the comparison results on the German Credit dataset in terms of the F1-score. It was observed that our proposed model consistently outperforms other baseline models regardless of the downstream model in use. The underlying driver for this observation is that \model\ has the capacity to customize the deep embedding space based on the performance evaluation conducted by a specific downstream machine learning model. This leads our model with a noteworthy degree of customization capability. 
Therefore, this experiment validates the robustness of our proposed model when confronted with different downstream tasks.

\begin{figure}[!t]
    \centering
    \includegraphics[width=\linewidth]{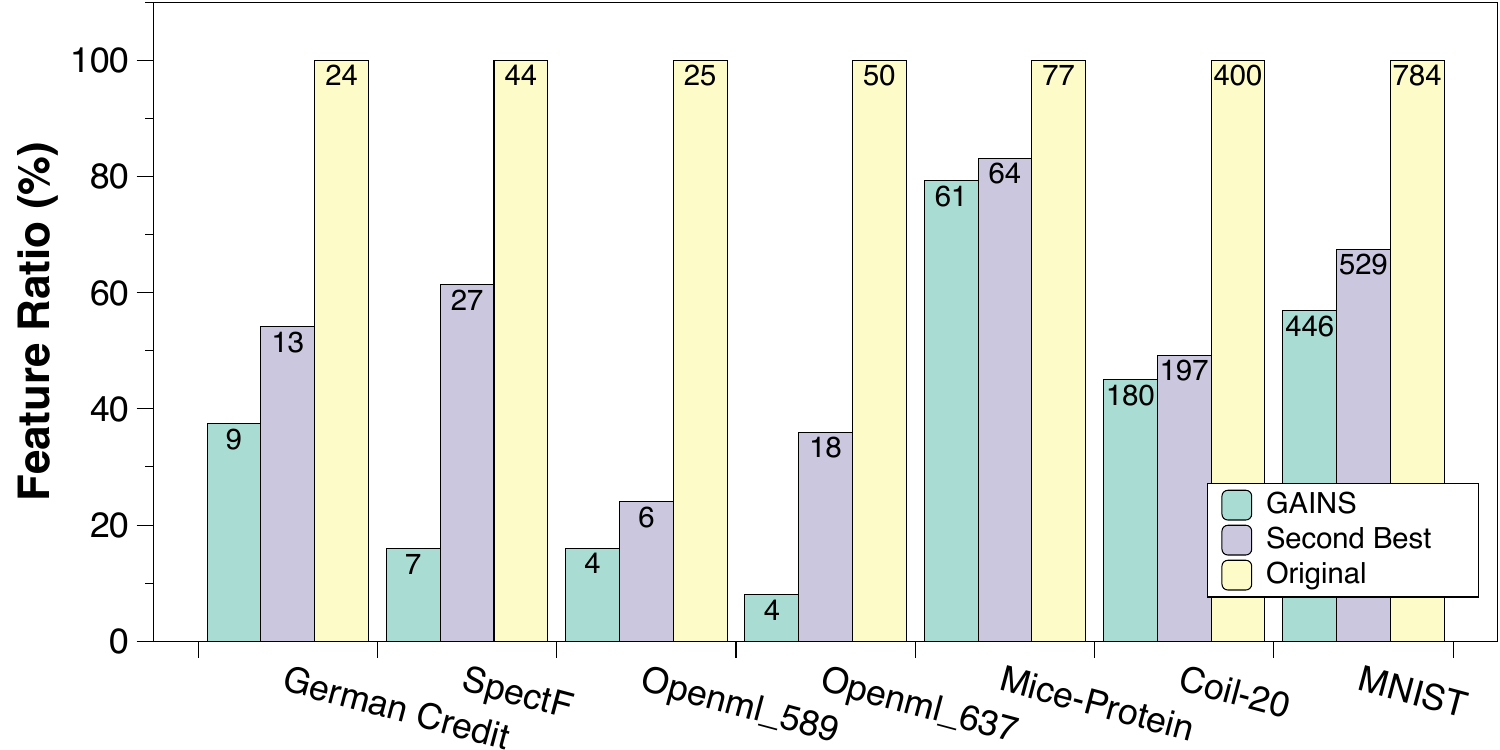}
    \caption{Comparison of the feature set size of  \model,  the best baseline model (Second Best), and the original feature set (Original).}
    \label{feat_num}
    \end{figure}

\smallskip
\noindent{\bf Study of the size of selected feature subsets.}
This experiment aims to answer this question: \textit{Is our proposed model capable of selecting a small, yet effective, feature subset?} We randomly selected seven datasets and compared the size of the feature set of our proposed model with the best-performing baseline model and the original feature set.
Figure~\ref{feat_num} shows the comparative results, represented in terms of feature ratio. The feature ratio provides an indication of the proportion of selected features relative to the entire feature set. 
We found that the feature subset selected by our model is substantially smaller than that of the second-best baseline, yet it maintains superior performance.
A plausible explanation for this observation is that the joint optimization of feature subset reconstruction loss and accuracy estimation loss enhances the denoising capability of gradient search, reduces noise and redundancy in features, and ultimately selects a compact but effective feature subset.
Therefore, this experiment demonstrates that the features selected by \model\ can not only improve model performance but also decrease computational expenses.

\begin{figure}[!t]
        \centering
        \subfigure[Step Size]{
        \includegraphics[width=4.25cm]{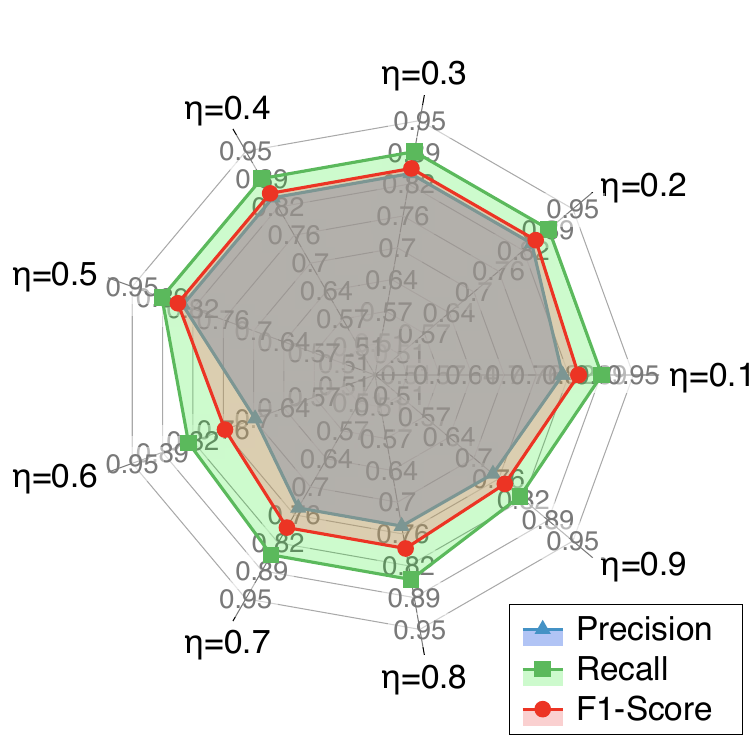}
        }
        \hspace{-3mm}
        \subfigure[Trade Off]{
        \includegraphics[width=4.25cm]{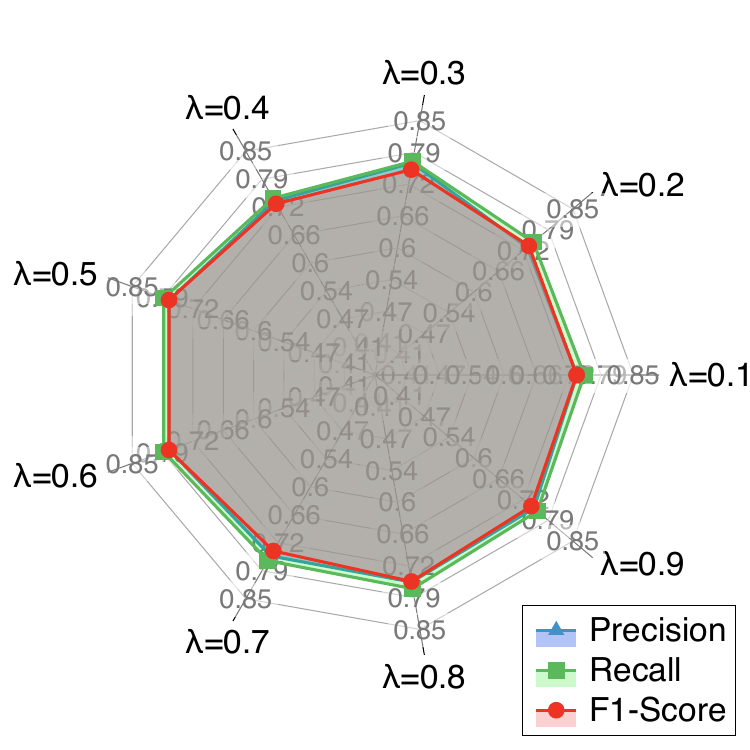}
        }
        \hspace{-3mm}
        \caption{The hyperparameter sensitivity test on German Credit.}
        \label{trade}
\end{figure}

\smallskip
\noindent{\bf Study of the hyperparameter sensitivity of \model.}
This experiment aims to answer: \textit{To what extent is the performance of \model\ sensitive to the values of the trade-off parameter $\lambda$ and step size $\eta$?} There are two major hyperparameters, training trade-off $\lambda$ and the step size $\eta$ of each search step. A higher $\lambda$ will make the model more concentrated on the loss from the reconstruction of sequence, and a higher $\eta$ make the model search more aggressive. We set $\eta$ varies from 0.1 to 1.0, and $\lambda$ from 0.1 to 0.9, then train the \model\ on German Credit. The model performance is reported in Figure~\ref{trade}. Overall, we observed that the model performance will change slightly on different step size settings. Further, a balanced (i.e., 0.5 or 0.6) model training trade-off value will slightly bring a higher downstream performance. These findings provide insight into how $\eta$ and $\lambda$ impact the performance of our model and how we might choose optimal values for these hyperparameters.

\section{Related Work}
\noindent\textbf{Feature Selection} can be broadly categorized as wrapper, filter, and embedded methods according to the selection strategies~\cite{li2017feature}.
Filter methods choose the highest-scoring features based on the feature relevance score derived from the statistical properties of the data~\cite{mrmr, biesiada2008feature, ding2014identification}.
For instance, K-Best~\cite{kbest} algorithm operates by ranking all features based on a specific criterion or metric. This criterion could be a correlation with the target variable, mutual information, or any other statistical measure that reflects the significance or importance of the feature. Once the features are ranked, the top "k" features are selected.
These approaches have low computational complexity and can efficiently select features from high-dimensional datasets.
But, they ignore feature-feature dependencies and interactions, resulting in suboptimal performance.
Wrapper methods assess the quality of the selected feature subset based on a predefined machine learning (ML) model in an iterative manner~\cite{gfe,sarlfs, fan2020autofs, fan2021autogfs, altarabichi2023fast}.
The performance of these methods is typically superior to that of filter methods because they evaluate the entire feature set.
MARLFS~\cite{marlfs} adopted a multi-agent reinforcement approach to manipulate and construct the combination of the features, which is effective to search the solution. 
Besides, many reinforcement learning approaches~\cite{wang2022group, xiao2022traceable, xiao2023tkdd} show great potential for feature engineering knowledge exploring.
However, enumerating all possible feature subsets is an NP-hard problem, leading to cannot identify the optimal feature subset.
Embedded methods convert the feature selection task into a regularization item in a prediction loss of the ML model to accelerate the selection process~\cite{lasso,lassonet,rfe,kumagai2022few,koyama2022effective}.
LassoNet~\cite{lassonet} is a representative embedded method that combines the regularization properties of the Lasso and the expressive power of neural networks. The method is designed to handle high-dimensional data with intricate and non-linear relationships among the features. The key principle behind LassoNet is that it introduces sparsity, just like Lasso, but in the weights of a neural network. LassoNet, thus, combines Lasso (Least Absolute Shrinkage and Selection Operator) with a feedforward neural network.
These methods may have outstanding performance on the incorporated ML model but are typically difficult to generalize to others.
Moreover, other works have proposed hybrid feature selection methods, which have two technical categories: 1) homogeneous approach~\cite{seijo2017testing,pes2017exploiting}; 2) heterogeneous approach~\cite{haque2016heterogeneous,seijo2019developing}.
Their performance is all limited by their basic aggregating methods.
Unlike these works, \model\ proposes a new feature selection perspective, which maps historical discrete selection records into a continuous embedding space and then employs the gradient-based search to 
 efficiently identify the optimal feature subset.

\section{Conclusion}
In conclusion, our research has challenged traditional views of feature selection as a discrete choice problem and instead considered a novel research question: is it feasible to approach feature selection within a continuous space, thus enhancing its automation, effectiveness, and generalizability?
In pursuit of an answer, we proposed a fresh perspective that transforms the discrete feature selection process into a gradient-optimized search, thereby reformulating feature selection as a continuous optimization task. 
Our study proposed a robust four-stage framework that integrates:
1) Automated reinforcement training data preparation,
2) Deep feature subset embedding,
3) Gradient-optimized feature subset search, and
4) Feature subset reconstruction.
Our research findings have offered significant insights:
1) We demonstrated that reinforcement feature selection can act as a powerful tool for automated training data collection, significantly enriching the diversity, scale, and comprehensiveness of the training data.
2) We established that the joint optimization of the encoder, evaluator, and decoder can efficiently construct a continuous embedding representation space.
3) We found that treating discrete feature selection as a gradient search strategy can effectively reduce feature subset sizes and boost generalization.
This novel method proved to be automated, effective, and input dimensionality agnostic.
Ultimately, these findings underscore the potential of our proposed approach in improving the effectiveness and efficiency of automated feature selection, thereby opening up new avenues for future research in this domain.

\section{Acknowledgement}
This work is partially supported by IIS-2152030, IIS-2045567, and IIS-2006889.


\bibliographystyle{IEEEtran}   
\bibliography{icdm}

\begin{thebibliography}{10}
\providecommand{\url}[1]{#1}
\csname url@samestyle\endcsname
\providecommand{\newblock}{\relax}
\providecommand{\bibinfo}[2]{#2}
\providecommand{\BIBentrySTDinterwordspacing}{\spaceskip=0pt\relax}
\providecommand{\BIBentryALTinterwordstretchfactor}{4}
\providecommand{\BIBentryALTinterwordspacing}{\spaceskip=\fontdimen2\font plus
\BIBentryALTinterwordstretchfactor\fontdimen3\font minus
  \fontdimen4\font\relax}
\providecommand{\BIBforeignlanguage}[2]{{%
\expandafter\ifx\csname l@#1\endcsname\relax
\typeout{** WARNING: IEEEtran.bst: No hyphenation pattern has been}%
\typeout{** loaded for the language `#1'. Using the pattern for}%
\typeout{** the default language instead.}%
\else
\language=\csname l@#1\endcsname
\fi
#2}}
\providecommand{\BIBdecl}{\relax}
\BIBdecl

\bibitem{kbest}
Y.~Yang and J.~O. Pedersen, ``A comparative study on feature selection in text
  categorization,'' in \emph{Icml}, vol.~97, no. 412-420.\hskip 1em plus 0.5em
  minus 0.4em\relax Nashville, TN, USA, 1997, p.~35.

\bibitem{forman2003extensive}
G.~Forman \emph{et~al.}, ``An extensive empirical study of feature selection
  metrics for text classification.'' \emph{J. Mach. Learn. Res.}, vol.~3, no.
  Mar, pp. 1289--1305, 2003.

\bibitem{hall1999feature}
M.~A. Hall, ``Feature selection for discrete and numeric class machine
  learning,'' 1999.

\bibitem{yu2003feature}
L.~Yu and H.~Liu, ``Feature selection for high-dimensional data: A fast
  correlation-based filter solution,'' in \emph{Proceedings of the 20th
  international conference on machine learning (ICML-03)}, 2003, pp. 856--863.

\bibitem{yang1998feature}
J.~Yang and V.~Honavar, ``Feature subset selection using a genetic algorithm,''
  in \emph{Feature extraction, construction and selection}.\hskip 1em plus
  0.5em minus 0.4em\relax Springer, 1998, pp. 117--136.

\bibitem{kim2000feature}
Y.~Kim, W.~N. Street, and F.~Menczer, ``Feature selection in unsupervised
  learning via evolutionary search,'' in \emph{Proceedings of the sixth ACM
  SIGKDD international conference on Knowledge discovery and data mining},
  2000, pp. 365--369.

\bibitem{narendra1977branch}
P.~M. Narendra and K.~Fukunaga, ``A branch and bound algorithm for feature
  subset selection,'' \emph{IEEE Transactions on computers}, no.~9, pp.
  917--922, 1977.

\bibitem{kohavi1997wrappers}
R.~Kohavi and G.~H. John, ``Wrappers for feature subset selection,''
  \emph{Artificial intelligence}, vol.~97, no. 1-2, pp. 273--324, 1997.

\bibitem{lasso}
R.~Tibshirani, ``Regression shrinkage and selection via the lasso,''
  \emph{Journal of the Royal Statistical Society: Series B (Methodological)},
  vol.~58, no.~1, pp. 267--288, 1996.

\bibitem{sugumaran2007feature}
V.~Sugumaran, V.~Muralidharan, and K.~Ramachandran, ``Feature selection using
  decision tree and classification through proximal support vector machine for
  fault diagnostics of roller bearing,'' \emph{Mechanical systems and signal
  processing}, vol.~21, no.~2, pp. 930--942, 2007.

\bibitem{marlfs}
K.~Liu, Y.~Fu, P.~Wang, L.~Wu, R.~Bo, and X.~Li, ``Automating feature subspace
  exploration via multi-agent reinforcement learning,'' in \emph{Proceedings of
  the 25th ACM SIGKDD International Conference on Knowledge Discovery \& Data
  Mining}, 2019, pp. 207--215.

\bibitem{lstm}
S.~Hochreiter and J.~Schmidhuber, ``Long short-term memory,'' \emph{Neural
  computation}, vol.~9, no.~8, pp. 1735--1780, 1997.

\bibitem{attention}
D.~Bahdanau, K.~Cho, and Y.~Bengio, ``Neural machine translation by jointly
  learning to align and translate,'' \emph{arXiv preprint arXiv:1409.0473},
  2014.

\bibitem{mrmr}
H.~Peng, F.~Long, and C.~Ding, ``Feature selection based on mutual information
  criteria of max-dependency, max-relevance, and min-redundancy,'' \emph{IEEE
  Transactions on pattern analysis and machine intelligence}, vol.~27, no.~8,
  pp. 1226--1238, 2005.

\bibitem{rfe}
P.~M. Granitto, C.~Furlanello, F.~Biasioli, and F.~Gasperi, ``Recursive feature
  elimination with random forest for ptr-ms analysis of agroindustrial
  products,'' \emph{Chemometrics and intelligent laboratory systems}, vol.~83,
  no.~2, pp. 83--90, 2006.

\bibitem{lassonet}
I.~Lemhadri, F.~Ruan, and R.~Tibshirani, ``Lassonet: Neural networks with
  feature sparsity,'' in \emph{International Conference on Artificial
  Intelligence and Statistics}.\hskip 1em plus 0.5em minus 0.4em\relax PMLR,
  2021, pp. 10--18.

\bibitem{gfe}
R.~Leardi, ``Genetic algorithms in feature selection,'' in \emph{Genetic
  algorithms in molecular modeling}.\hskip 1em plus 0.5em minus 0.4em\relax
  Elsevier, 1996, pp. 67--86.

\bibitem{sarlfs}
K.~Liu, P.~Wang, D.~Wang, W.~Du, D.~O. Wu, and Y.~Fu, ``Efficient reinforced
  feature selection via early stopping traverse strategy,'' in \emph{2021 IEEE
  International Conference on Data Mining (ICDM)}.\hskip 1em plus 0.5em minus
  0.4em\relax IEEE, 2021, pp. 399--408.

\bibitem{seijo2017ensemble}
B.~Seijo-Pardo, I.~Porto-D{\'\i}az, V.~Bol{\'o}n-Canedo, and
  A.~Alonso-Betanzos, ``Ensemble feature selection: homogeneous and
  heterogeneous approaches,'' \emph{Knowledge-Based Systems}, vol. 118, pp.
  124--139, 2017.

\bibitem{mcdm}
A.~Hashemi, M.~B. Dowlatshahi, and H.~Nezamabadi-pour, ``Ensemble of feature
  selection algorithms: a multi-criteria decision-making approach,''
  \emph{International Journal of Machine Learning and Cybernetics}, vol.~13,
  no.~1, pp. 49--69, 2022.

\bibitem{paszke2019pytorch}
A.~Paszke, S.~Gross, F.~Massa, A.~Lerer, J.~Bradbury, G.~Chanan, T.~Killeen,
  Z.~Lin, N.~Gimelshein, L.~Antiga \emph{et~al.}, ``Pytorch: An imperative
  style, high-performance deep learning library,'' \emph{Advances in neural
  information processing systems}, vol.~32, 2019.

\bibitem{li2017feature}
J.~Li, K.~Cheng, S.~Wang, F.~Morstatter, R.~P. Trevino, J.~Tang, and H.~Liu,
  ``Feature selection: A data perspective,'' \emph{ACM Computing Surveys
  (CSUR)}, vol.~50, no.~6, pp. 1--45, 2017.

\bibitem{biesiada2008feature}
J.~Biesiada and W.~Duch, ``Feature selection for high-dimensional data—a
  pearson redundancy based filter,'' in \emph{Computer recognition systems
  2}.\hskip 1em plus 0.5em minus 0.4em\relax Springer, 2008, pp. 242--249.

\bibitem{ding2014identification}
H.~Ding, P.-M. Feng, W.~Chen, and H.~Lin, ``Identification of bacteriophage
  virion proteins by the anova feature selection and analysis,''
  \emph{Molecular BioSystems}, vol.~10, no.~8, pp. 2229--2235, 2014.

\bibitem{fan2020autofs}
W.~Fan, K.~Liu, H.~Liu, P.~Wang, Y.~Ge, and Y.~Fu, ``Autofs: Automated feature
  selection via diversity-aware interactive reinforcement learning,'' in
  \emph{2020 IEEE International Conference on Data Mining (ICDM)}.\hskip 1em
  plus 0.5em minus 0.4em\relax IEEE, 2020, pp. 1008--1013.

\bibitem{fan2021autogfs}
W.~Fan, K.~Liu, H.~Liu, A.~Hariri, D.~Dou, and Y.~Fu, ``Autogfs: Automated
  group-based feature selection via interactive reinforcement learning,'' in
  \emph{Proceedings of the 2021 SIAM International Conference on Data Mining
  (SDM)}.\hskip 1em plus 0.5em minus 0.4em\relax SIAM, 2021, pp. 342--350.

\bibitem{altarabichi2023fast}
M.~G. Altarabichi, S.~Nowaczyk, S.~Pashami, and P.~S. Mashhadi, ``Fast genetic
  algorithm for feature selection—a qualitative approximation approach,''
  \emph{Expert systems with applications}, vol. 211, p. 118528, 2023.

\bibitem{wang2022group}
D.~Wang, Y.~Fu, K.~Liu, X.~Li, and Y.~Solihin, ``Group-wise reinforcement
  feature generation for optimal and explainable representation space
  reconstruction,'' \emph{Proceedings of the 28th ACM SIGKDD international
  conference on Knowledge discovery and data mining}, 2022.

\bibitem{xiao2022traceable}
M.~Xiao, D.~Wang, M.~Wu, Z.~Qiao, P.~Wang, K.~Liu, Y.~Zhou, and Y.~Fu,
  ``Traceable automatic feature transformation via cascading actor-critic
  agents,'' in \emph{Proceedings of the 2023 SIAM International Conference on
  Data Mining (SDM)}.\hskip 1em plus 0.5em minus 0.4em\relax SIAM, 2023, pp.
  775--783.

\bibitem{xiao2023tkdd}
M.~Xiao, D.~Wang, M.~Wu, K.~Liu, H.~Xiong, Y.~Zhou, and Y.~Fu, ``Traceable
  group-wise self-optimizing feature transformation learning: A dual
  optimization perspective,'' 2023.

\bibitem{kumagai2022few}
A.~Kumagai, T.~Iwata, Y.~Ida, and Y.~Fujiwara, ``Few-shot learning for feature
  selection with hilbert-schmidt independence criterion,'' \emph{Advances in
  Neural Information Processing Systems}, vol.~35, pp. 9577--9590, 2022.

\bibitem{koyama2022effective}
K.~Koyama, K.~Kiritoshi, T.~Okawachi, and T.~Izumitani, ``Effective nonlinear
  feature selection method based on hsic lasso and with variational
  inference,'' in \emph{International Conference on Artificial Intelligence and
  Statistics}.\hskip 1em plus 0.5em minus 0.4em\relax PMLR, 2022, pp.
  10\,407--10\,421.

\bibitem{seijo2017testing}
B.~Seijo-Pardo, V.~Bol{\'o}n-Canedo, and A.~Alonso-Betanzos, ``Testing
  different ensemble configurations for feature selection,'' \emph{Neural
  Processing Letters}, vol.~46, no.~3, pp. 857--880, 2017.

\bibitem{pes2017exploiting}
B.~Pes, N.~Dess{\`\i}, and M.~Angioni, ``Exploiting the ensemble paradigm for
  stable feature selection: a case study on high-dimensional genomic data,''
  \emph{Information Fusion}, vol.~35, pp. 132--147, 2017.

\bibitem{haque2016heterogeneous}
M.~N. Haque, N.~Noman, R.~Berretta, and P.~Moscato, ``Heterogeneous ensemble
  combination search using genetic algorithm for class imbalanced data
  classification,'' \emph{PloS one}, vol.~11, no.~1, p. e0146116, 2016.

\bibitem{seijo2019developing}
B.~Seijo-Pardo, V.~Bol{\'o}n-Canedo, and A.~Alonso-Betanzos, ``On developing an
  automatic threshold applied to feature selection ensembles,''
  \emph{Information Fusion}, vol.~45, pp. 227--245, 2019.

\end{thebibliography}

\end{document}